\documentclass[12pt, a4paper]{article}
\pdfoutput=1
\usepackage[utf8]{inputenc}
\usepackage{amsmath}
\usepackage{amssymb}
\usepackage{subcaption}
\usepackage{booktabs}
\usepackage{graphicx}
\usepackage[colorlinks,
            linkcolor=blue,
            anchorcolor=blue,
            citecolor=blue]{hyperref} 
\usepackage{cleveref}
\usepackage{multirow}

\usepackage[T1]{fontenc}
\usepackage[utf8]{inputenc}
\usepackage{authblk}
\crefname{figure}{Figure.}{Figures.}
\crefname{table}{Table.}{Tables.}
\crefname{section}{Section.}{Sections.}

\title{TriangleNet: Edge Prior Augmented Network for Semantic Segmentation through Cross-Task Consistency}
\author[1]{\small Dan Zhang}
\author[1]{Rui Zheng \thanks{Corresponding author}}
\author[2]{Luosang Gadeng}
\author[3]{Pei Yang}
\affil[1]{School of Information and Engineering, Minzu University of China, Beijing 100081, China}
\affil[2]{Department of Information Science and Technology, Tibet University, Lhasa 850012, China}
\affil[3]{Department of Computer Technology and Application, Qinghai University, Xining 810016, China}
\affil[ ]{zhangdan@muc.edu.cn, rzhengbj@163.com, lsgd@utibet.edu.cn, yangpeinmgdx@sina.com}

\date{}

\begin{document}
\maketitle
\begin{abstract}

This paper addresses the task of semantic segmentation in computer vision, aiming to achieve precise pixel-wise classification. We investigate the joint training of models for semantic edge detection and semantic segmentation, which has shown promise. However, implicit cross-task consistency learning in multi-task networks is limited. To address this, we propose a novel "decoupled cross-task consistency loss" that explicitly enhances cross-task consistency. Our semantic segmentation network, TriangleNet, achieves a substantial 2.88\% improvement over the Baseline in mean Intersection over Union (mIoU) on the Cityscapes test set. Notably, TriangleNet operates at 77.4\% mIoU/46.2 FPS on Cityscapes, showcasing real-time inference capabilities at full resolution. With multi-scale inference, performance is further enhanced to 77.8\%. Furthermore, TriangleNet consistently outperforms the Baseline on the FloodNet dataset, demonstrating its robust generalization capabilities. The proposed method underscores the significance of multi-task learning and explicit cross-task consistency enhancement for advancing semantic segmentation and highlights the potential of multitasking in real-time semantic segmentation.

\end{abstract}

\section{Introduction}
The combination of image semantic segmentation and deep learning has gone through a long period of time, accumulating a large number of excellent works such as FCN \cite{long2015fully}, U-Net \cite{ronneberger2015u}, FastFCN \cite{wu2019fastfcn}, Gated-SCNN \cite{takikawa2019gated}, DeepLab Series \cite{chen2014semantic,chen2017deeplab,chen2017rethinking}, Mask R-CNN \cite{he2017mask} and so on, as well as leaving unsolved problems. The main challenge is the fine-grained localization of pixel labels \cite{ulku2022survey}. The prevailing structure of semantic segmentation networks mostly follows the encoder-decoder structure adopted by FCN \cite{long2015fully}. First, downsampling is used to expand the receptive field to extract high-level semantics, and then upsampling is used to recover low-level details. The edge details lost by conventional downsampling operations in semantic segmentation networks are difficult to recover during upsampling. A compensatory solution is to introduce additional knowledge among which edge priors are intuitive and easily accessible. In order to inject edge priors into semantic segmentation networks, one way is to train a semantic edge detection model and a semantic segmentation model jointly. General practice is a two-stream framework that trains a semantic edge detection branch and a semantic segmentation branch in a hard parameter-sharing manner \cite{ruder2017overview}. The predictions of the semantic edge detection branch on edge points may differ from those of the semantic segmentation branch, which implies the existence of cross-task inconsistency. Conventionally, a fusion module is introduced to cope with this conflict, such as \cite{zhen2020joint,liu2020auxiliary} do, which intends to fuse features from the semantic edge detection branch to improve the semantic segmentation branch. However, the effects of these fusion modules are sometimes not as effective as expected. As the ablation experiments of \cite{zhen2020joint} point out, the improvement of the mean of class-wise intersection-over-union (mIoU) on the Cityscapes validation set mainly depends on duality loss(+1.44\%) rather than semantic edge fusion(+0.22\%), or pyramid context module(+0.62\%). A considerable amount of segmentation errors along object boundaries still exist, which means the mutual consistency between the semantic segmentation branch and the semantic edge detection branch should be further studied to improve the quality of segmentation results.

\begin{figure}[ht]
    \centering
    \includegraphics[width=0.8\linewidth]{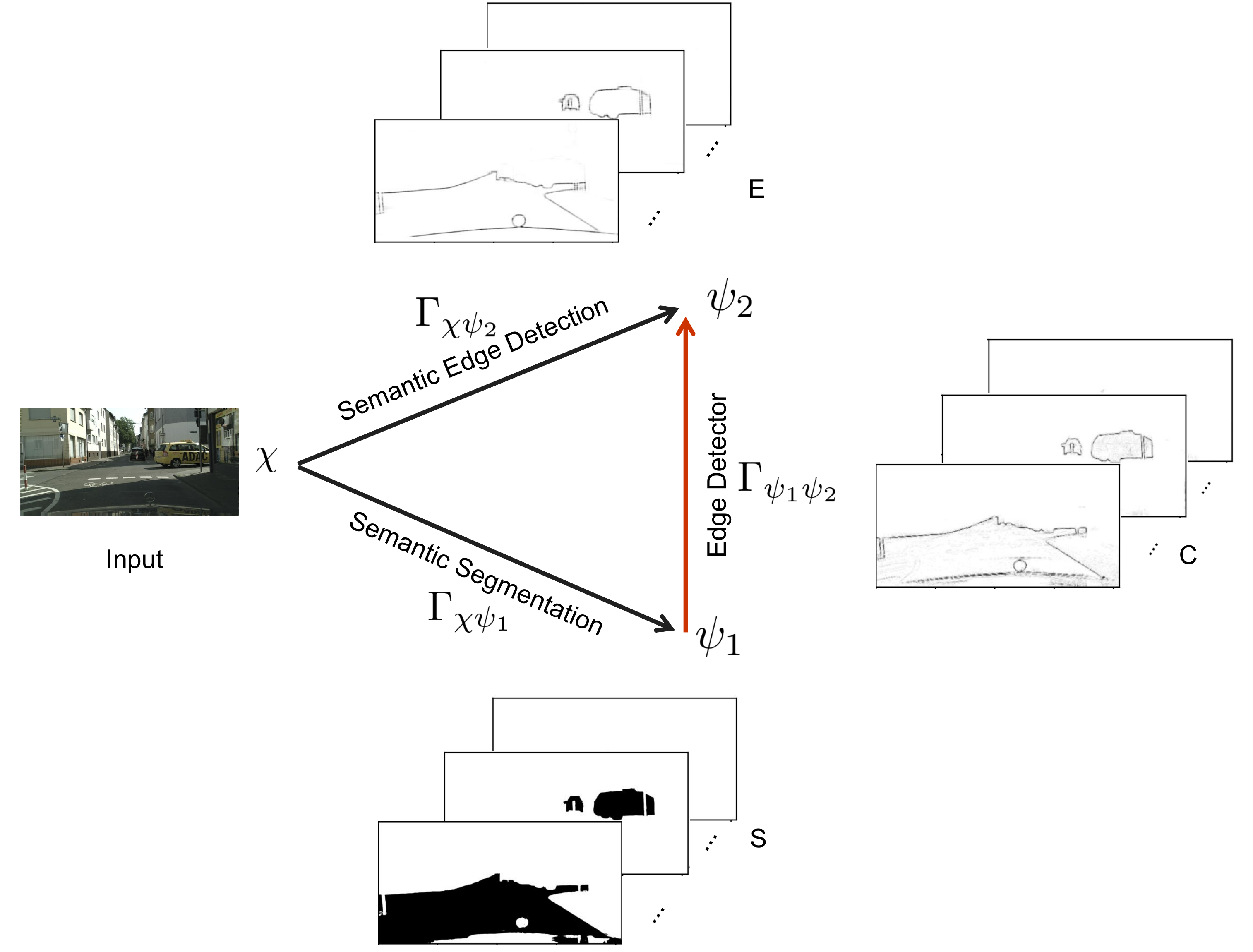}
    \caption{The multi-task learning framework of semantic segmentation and semantic edge detection coincides with the Elementary Consistency Unit theory where the prediction $\chi \rightarrow \psi_{1}$ is enforced to be consistent with $\chi \rightarrow \psi_{2}$ using a function that relates $\psi_{1}$ to $\psi_{2}$. $S$, $E$, and $C$ respectively denote the outputs processed through $\Gamma_{\chi\psi_{1}}$, $\Gamma_{\chi\psi_{2}}$, and $\Gamma_{\psi_{1}\psi_{2}}$.}
    \label{Figure:theory}
\end{figure}

We have observed that many semantic segmentation works can be loosely viewed as semantic edge detection tasks, since applying edge detectors to semantic segmentation outputs can yield semantic edge results. Their relationship can be modeled as \cref{Figure:theory}. Logically, in order to conserve consistency among tasks, the results of inferring semantic edges from an input image should be the same regardless of the inference paths. That is, predicting semantic edges by first predicting semantic segmentation maps from an input image should achieve similar predictions as directly predicting semantic edges from the input image. This observation aligns with the concept of inference-path invariance, which serves as the guiding ideology in the work by Zamir et al. \cite{zamir2020robust}. The concept emphasizes that predictions should remain consistent regardless of the specific inference paths. The input image domain, the semantic segmentation domain, and the semantic edge domain form an Elementary Consistency Unit proposed by Zamir et al. \cite{zamir2020robust}, which can be illustrated by \cref{Figure:theory}. 

By imposing a cross-task consistency loss on the endpoint outputs of the two paths, the consistency between semantic segmentation and semantic edge detection can be explicitly learned. Based on these analyses, we propose a new framework to simultaneously train a semantic segmentation branch and a semantic edge detection branch, and the overall process is shown in \cref{Figure:pipeline}.

The highlights of this paper are as follows.

\begin{figure}[ht]
    \centering
    \includegraphics[width=0.95\linewidth]{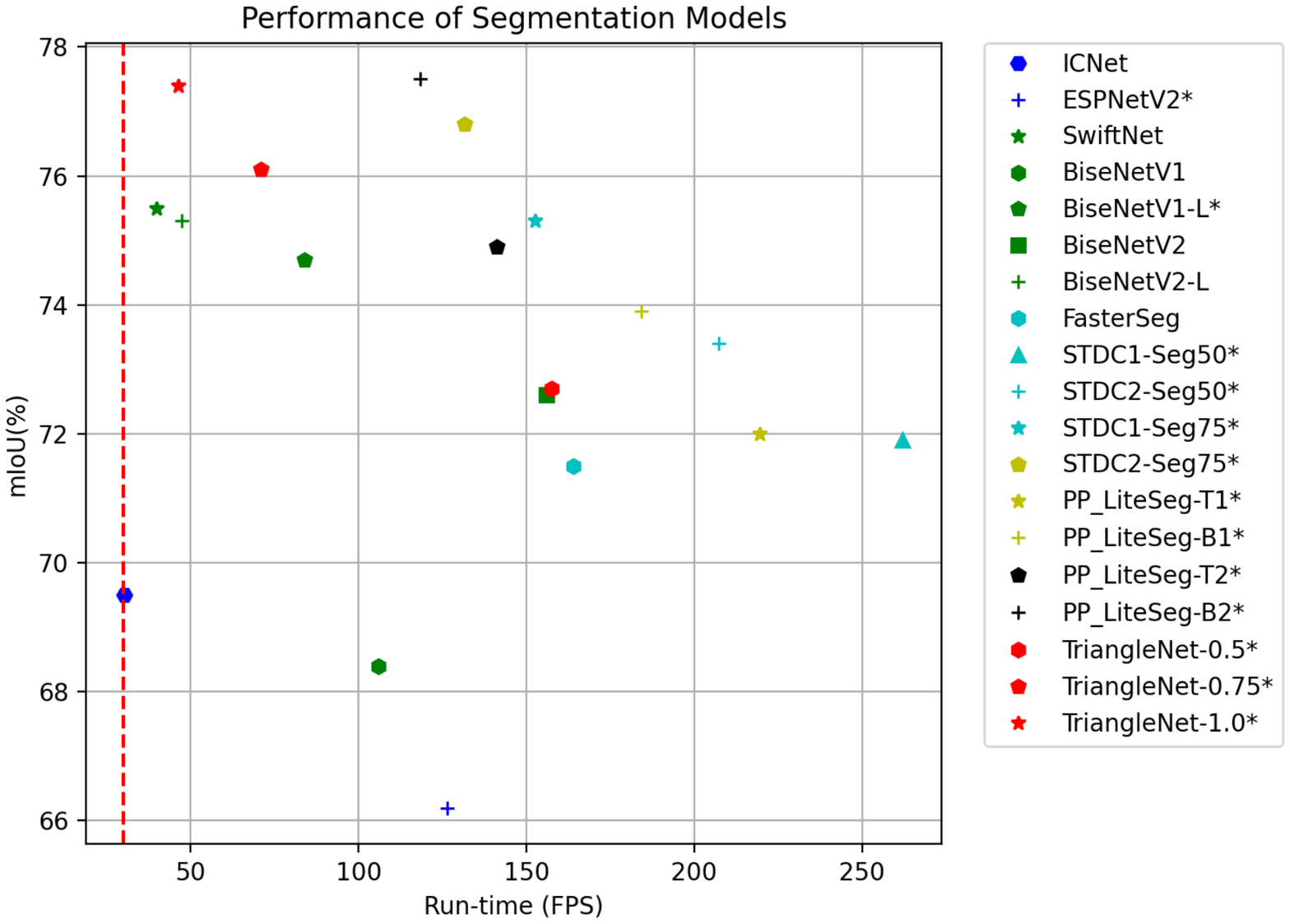}
    \caption{Run-time/accuracy trade-off comparison on the Cityscapes test set. Our models (in red) achieves an excellent run-time vs. accuracy trade-off among all previous real-time methods. FPS=30 is the red line dividing real-time and non-real-time performance in the graph. The asterisk after the model name indicates that the inference speeds of these models were obtained using the same deep learning framework, PaddlePaddle, and the same hardware platform, A100 40G device.}
    \label{Figure:fps}
\end{figure}

\begin{enumerate}

\item Figure \ref{Figure:fps} illustrates the superior balance between speed and accuracy achieved by our framework on the Cityscapes dataset, distinguishing it as one of the few models capable of real-time inference at full resolution. Notably, our model operates at an impressive 77.4\% mIoU while maintaining a fast frame rate of 46.2 FPS on Cityscapes.

\item We introduce a novel approach, "decoupled cross-task consistency loss," to explicitly enhance cross-task consistency between semantic edge detection and semantic segmentation, resulting in 1.83\% improvement in mIoU on the Cityscapes test set. The decoupled loss effectively enforces consistency across tasks, facilitating the learning of shared representations and leading to improved overall performance.

\item Our model demonstrates exceptional efficacy in categories characterized by distinct edges and boundaries, as evidenced by some categories achieving significant IoU improvements, with "train" nearly reaching an 18\% increase in IoU on the Cityscapes test set. These results further reinforce the importance of incorporating edge information through our approach, highlighting its impact on enhancing segmentation performance.

\item The decoupled architecture we have designed allows for joint training of multiple tasks without the need for fusion modules during inference, thereby avoiding the introduction of extra inference overhead. This efficient and practical approach enables us to leverage the advantages of multitasking for real-time semantic segmentation without compromising on performance.

\end{enumerate}

\section{Related Work}
\subsection{Semantic Segmentation}
Strengths, weaknesses, and major challenges of semantic segmentation are extensively discussed in the literature \cite{hao2020brief, vandenhende2021multi, minaee2021image, ulku2022survey}. There are currently two approaches to semantic segmentation: improving the object's inner consistency or refining details along objects' boundaries.

The inner inconsistency of the object is attributed to the limited receptive field, by which the longer-range relationships of pixels in an image cannot be fully modeled. Consequently, dilated convolution \cite{yu2017dilated} or high-resolution network \cite{sun2019deep} is introduced to enlarge the receptive field. Furthermore, many attempts have been made to capture contextual information, such as recurrent networks \cite{pinheiro2014recurrent,byeon2015scene}, pyramid pooling module \cite{zhao2017pyramid}, graph convolution networks \cite{zhang2019dual}, CRF related networks \cite{chen2014semantic,chen2017deeplab,zheng2015conditional}, non-local operator \cite{wang2018non}, attention mechanism \cite{fu2019dual, cheng2022mifnet}, etc.

The ambiguity along edges is caused by down-sampling operations in the FCNs that result in blurred predictions. It is difficult to recover spatial information lost during down-sampling through simple up-sampling. Thus, previous papers have made efforts to add priors to guide the upsampling process, many of which focus on the use of edge priors. The general practice is a two-stream framework that trains an edge detection branch and a semantic segmentation branch jointly, which will be elaborated later.

\subsection{Multi-Task Learning}
Driven by deep learning, many dense prediction tasks such as semantic segmentation, instance segmentation, etc. have achieved significant performance improvements. Typically, tasks are learned in isolation, i.e. each task is trained with a separate neural network. Recently, multi-task learning (MTL) techniques that learn shared representations by jointly processing multiple tasks have shown promising results.

Almost all theories about MTL are based on the assumption that tasks learned together should be relevant, or a phenomenon called negative transfer would occur. In practice, it is more dependent on expert experience to find relevant tasks. For example, \cite{zhang2018joint,mousavian2016joint,nekrasov2019real,he2021sosd} jointly train semantic segmentation and depth estimation to achieve better results, and \cite{de2018panoptic,zhao2020jsnet} jointly train semantic segmentation and instance segmentation to increase accuracy. \cite{takikawa2019gated,ding2019boundary,chen2016semantic,liu2020auxiliary,zhen2020joint} train semantic segmentation and edge detection jointly to improve metrics. Among these, the edge priors can be further subdivided into binary edge priors and semantic edge priors. For example, in GSCNN \cite{takikawa2019gated}, the binary edge is used as a gate to improve performance. In BFP \cite{ding2019boundary}, binary edge information is used to propagate local features within regions. \cite{chen2016semantic} adopts domain transform to perform edge-preserving filtering controlled by a binary edge map derived from a task-specific edge detection task. \cite{liu2020auxiliary} applies explicit semantic boundary supervision to learn semantic features and edge features in parallel and an attention-based feature fusion module to combine the high-resolution edge features with wide-receptive-field semantic features. RPCNet \cite{zhen2020joint} presents an interacting multi-task learning framework for semantic segmentation and semantic boundary detection.

The most common multi-task learning framework shares some layers in the feature extraction stage and designs independent layers for each specific task, which is called the hard parameter-sharing approach \cite{ruder2017overview}. This approach makes it difficult to ensure that multiple tasks can work together. Although there have been some means of using uncertainty \cite{kendall2018multi} to determine weights of tasks, the relationship between tasks is still not very clear, which drives the study of explicit consistency constraints between tasks.

\subsection{Consistency Learning}
It has been speculated that multi-task networks may automatically produce cross-task consistent predictions since their representations are shared. Numerous studies \cite{kokkinos2017ubernet,xu2018pad,standley2020tasks,zamir2020robust} have observed that this is not necessarily true, since consistent learning is not enforced directly during training, indicating the need for explicit enhancement of consistency during learning.

From the literature, two kinds of explicit consistency constraints can be summarized. One idea is formulated as the Cross-Task Consistency theory based on inference-path invariance by \cite{zamir2020robust}. \cite{zamir2020robust} first analyzes the Cross-Task Consistency theory of the triangular shown in \cref{Figure:theory}, and deduces the formula based on the $l_{1}$ norm assumption. Then it generalizes to cases where in the larger system of domains, consistency can be enforced using invariance along arbitrary paths, as long as their endpoints are the same. \cite{nakano2021cross} conveys the same insight, which uses the predictions of one task as inputs to another network to predict the other task, obtaining task-transferred predictions. Explicit constraints are imposed between the transferred prediction with the prediction of the other task.

Another idea is that for a specific geometric feature, such as the boundary, the results extracted by different tasks should be consistent. For instance, \cite{zhu2020edge} force the depth border to be consistent with the segmentation border through morphing. \cite{papadopoulos2021semantic} penalizes differences between the edges of the semantic heat map and the edges of the depth map through a holistic consistency loss.

\subsection{Real-Time Semantic Segmentation}

Real-time semantic segmentation is a challenging and essential task in computer vision, aiming to perform pixel-wise classification with high accuracy and rapid inference speed. The demand for real-time processing in applications such as autonomous vehicles, robotics, and augmented reality has driven extensive research efforts to develop efficient algorithms and architectures. Attempts to achieve a balance between speed and accuracy in real-time semantic segmentation include efficient architectures \cite{paszke2016enet, mehta2018espnet}, lightweight convolutions \cite{zhang2018shufflenet, sandler2018mobilenetv2}, knowledge distillation \cite{hinton2015distilling, liu2019structured, he2019knowledge}, pruning and quantization \cite{liu2017learning, jacob2018quantization}, and optimization techniques \cite{sun2017learning}.

Multi-task learning has shown promise in various computer vision tasks, but it is relatively less popular in real-time semantic segmentation. One of the challenges in deploying multi-task learning for real-time semantic segmentation is the potential increase in inference overhead due to fusion modules or additional processing steps. While multi-task learning can be beneficial during training by leveraging shared representations and learning complementary features from related tasks, the goal is to incorporate this knowledge effectively without introducing extra inference time. To achieve this, researchers are exploring methods to learn shared representations without the need for explicit fusion modules during inference. Some approaches to address this concern include decoupled architectures \cite{hong2015decoupled}, knowledge distillation, shared layers \cite{hu2020real}, weight sharing \cite{paszke2016enet, mehta2018espnet}.

\subsection{Edge Detection}
One notable method of applying deep neural networks to train and predict edges in an image-to-image fashion and end-to-end training is the holistically-nested edge detection (HED) \cite{xie2015holistically}. HED is a binary edge detection network, where the edge pixels are all set to 1 and 0 otherwise. Practically, edge pixels appear in contours or junctions belonging to two or more semantics, resulting in a challenging category-aware semantic edge detection problem. A pioneering approach is given by CASENet \cite{yu2017casenet} that extends the work of HED. However, both HED and CASENet employ fixed weight fusion to merge side outputs, ignoring image-specific and location-specific information. To 
address this, DFF \cite{hu2019dynamic} designed an adaptive weight fusion module to assign different fusion weights for different input images and locations adaptively.

In essence, compared to binary edge detection, semantic edge detection is more coupled with semantic segmentation since it provides semantic information about edge pixels while locating edges. 

\section{Method}
In this section, we will first introduce the overview pipeline of our architecture illustrated in \cref{Figure:pipeline} and then explain the components in detail.
\subsection{Model Overview}
\begin{figure}[ht]
  \centering
  \includegraphics[width=0.95\linewidth]{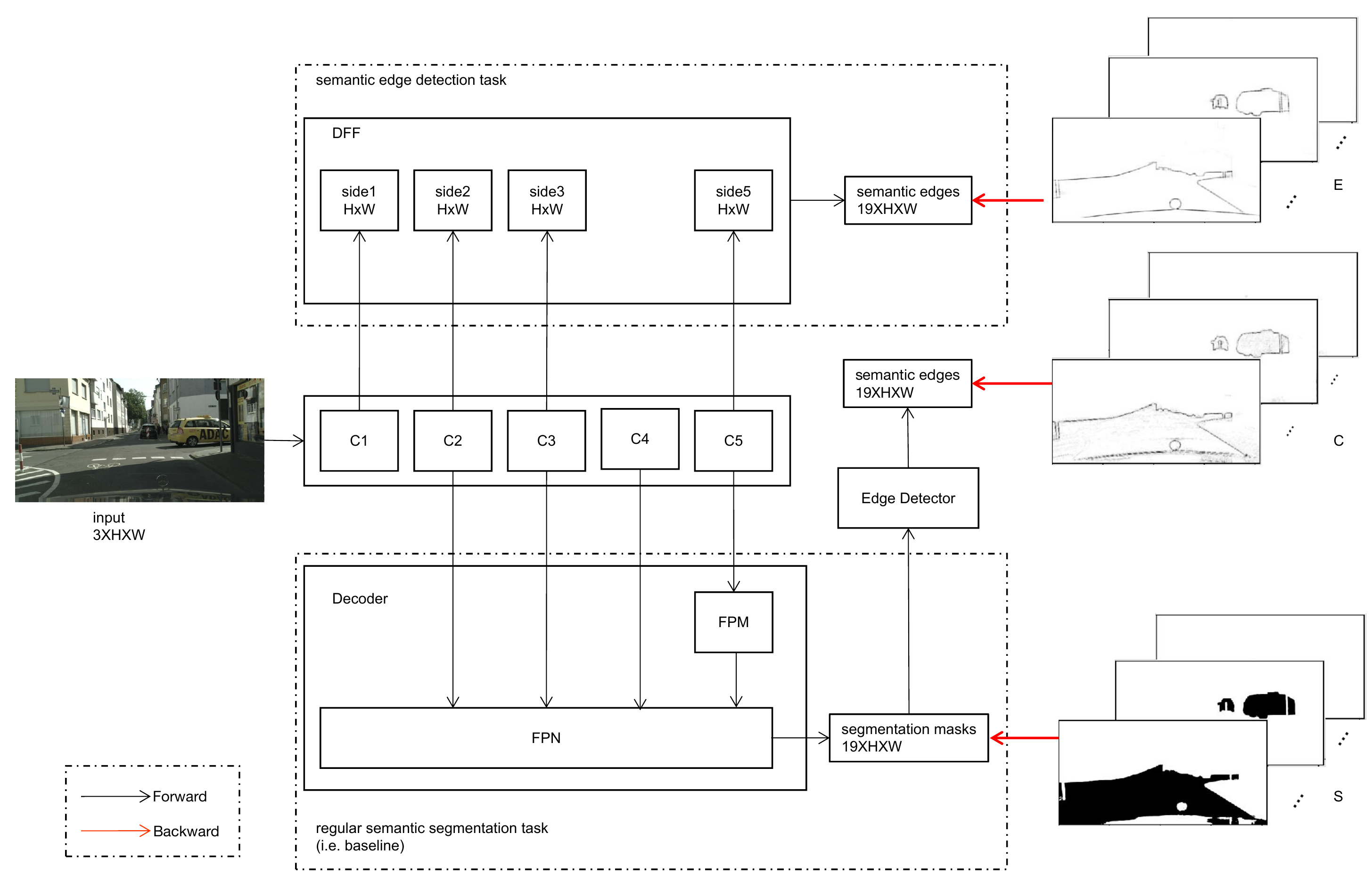}
  \caption{The overall pipeline of TriangleNet. The shared backbone network produces 5-layer features. The task-specific parts of the two branches are enclosed by dashed boxes.}
  \label{Figure:pipeline}
\end{figure}

As shown in \cref{Figure:pipeline}, the overall network is a two-stream framework following a hard parameter-sharing manner \cite{ruder2017overview}. It contains two branches; the upper one is an implementation of DFF \cite{hu2019dynamic} responsible for semantic edge detection, and the lower one is an FCN for semantic segmentation integrated with PPM \cite{zhao2017pyramid} and FPN \cite{lin2017feature} as the decoder. The backbone of the FCN is replaceable, the features output by which are shared by the two branches. Except for the shared backbone layers, the other layers of the two branches are task-specific and parallel. An Edge Detector is used to transfer the segmentation maps to semantic edges, which are enforced to be consistent with the output of the semantic edge detection branch by a consistency loss. We call this network TriangleNet because its underlying theory can be formulated by a triangular relation shown in \cref{Figure:theory}. 
\subsection{Edge Detector}
There are various strategies to extract semantic edges from the results of semantic segmentation. An edge detection operator such as Canny \cite{canny1986computational}, a Spatial Gradient Solution \cite{zhen2020joint} and a transfer network are optional solutions. To guarantee end-to-end training, the chosen scheme must be differentiable. For simplicity, we choose the Spatial Gradient Solution, which uses adaptive pooling to derive a spatial gradient. The formulation is as follows:
\begin{equation}
    \nabla S_{k}\left ( p \right ) =\mid S_{k}\left ( p \right ) -\operatorname{pool}_{w}\left (  S_{k}\left (  p\right ) \right )  \mid,
\label{formula:adaptive_pool}
\end{equation}
where $S$ represents the probability map of semantic segmentation, $S_{k}\left ( p \right )$ indicates the predicted probability on the $k$-th semantic category at pixel $p$, and \(\mid \cdot \mid\)  remarks the absolute value function. \(pool_{w}\) is an adaptive average pooling operation with kernel size $w$. The same as \cite{zhen2020joint}, $w$ is used to control the derived boundary width and is set to 3. 
\subsection{Task-specific Elementary Consistency Unit\label{sec:tsecn}}
TriangleNet consists of three domains: the input image domain, the semantic segmentation domain, and the semantic edge domain. As illustrated by \cref{Figure:theory}, $\chi$ denotes the query domain(e.g. input RGB images), \begin{math}\psi=\left \{ \psi_{1},\psi_{2} \right \} \end{math} is the set of two desired prediction domains. Specifically, $\psi_{1}$ represents the semantic segmentation domain and $\psi_{2}$ represents the semantic edge domain. The functions that map the query domain onto prediction domains are defined as $\Gamma _{\chi\psi_{i}}\left ( i=1,2 \right )$ which outputs $\psi_{i}$ given $\chi$. $\Gamma_{\psi_{1}\psi_{2}}$ denotes the cross-task function that maps the semantic segmentation domain to the semantic edge domain. According to the Elementary Consistency Unit theory proposed by \cite{zamir2020robust}, predicting $\psi_{2}$ by first predicting $\psi_{1}$ from $\chi$ should achieve predictions similar to directly predicting $\psi_{2}$ from $\chi$. 

To enhance the comprehension of the consistent constraint across the three domains, in \cref{Figure:theory}, we provided visual examples for each domain. $S$ represents an instance from domain $\psi_{1}$, while $E$ and $C$ are instances from domain $\psi_{2}$. Here, $E$ corresponds to the output of the semantic edge detection model, while $C$ is the output of the semantic segmentation model after undergoing the edge detector process. Notably, $E$ and $C$ exhibit a high degree of similarity, making them highly comparable and indicative of strong consistency between the two outputs.

\subsubsection{Ohem Cross Entropy Loss}
In our framework, $\Gamma_{\chi\psi_{i}}$ are neural networks. Through $\Gamma_{\chi\psi_{1}}$ we can obtain the semantic segmentation probability map \begin{math} S \in R^{H\times W\times K} \end{math}, where $K$ is the number of categories. A common way of training the neural network in $\Gamma_{\chi\psi_{1}}$  is to find parameters of $\Gamma_{\chi\psi_{1}}$ that minimize a loss called Cross Entropy Loss. 

What we actually use is an improved version called the Ohem Cross Entropy Loss implemented by PaddleSeg \cite{liu2021paddleseg}. It stands for "Online Hard Example Mining Cross Entropy Loss." Instead of considering the loss for all examples in a batch, it selects only the hard examples and uses those examples to update the model during training. This helps in dealing with class imbalance and emphasizing difficult examples that can lead to better generalization. Hard examples are considered those examples with low probabilities of the relevant label. In other words, they are examples that the model finds challenging to classify correctly. We denote Ohem Cross Entropy Loss as $L_{s}$, which measures the difference between S and the semantic segmentation ground truth.

The formula for Ohem Cross Entropy Loss can be expressed as follows: 

\begin{equation}
    L_{s}=- \frac{1}{N_{hard}}  {\textstyle \sum_{p\in hard\_examples}^{}}Y(p)logP(p),
\label{formula:Ls}
\end{equation}

where $Y(p)$ denotes the ground truth label at pixel $p$. $P(p)$ represents the probability of the corresponding label at pixel $p$.
$N_{hard}$ is the number of hard examples to be considered. It can be a fixed number or a percentage of the batch size, depending on implementation. In PaddleSeg \cite{liu2021paddleseg}, the number of hard examples is determined by the $min\_kept$ and $thresh$ hyperparameters, where $min\_kept$ specifies the minimum number of hard examples to be kept, and $thresh$ sets the probability threshold below which examples are considered hard.

\subsubsection{Semantic Edge Loss}
Through $\Gamma_{\chi\psi_{2}}$ we can obtain the semantic edge probability map \begin{math} E \in R^{H\times W\times K} \end{math}. While training the neural network in $\Gamma_{\chi\psi_{2}}$, we minimize a loss called Multi-Label Loss, which is formulated as: 
\begin{equation}
    L_{e}=-\sum_{k} \sum_{p}  G_{k}\left (  p\right )  \log E_{k}\left (  p\right )+\left(1-G_{k}\left (  p\right )\right) \log \left(1-E_{k}\left (  p\right )\right),
\label{formula:Le}
\end{equation}
where $G_{k}\left (  p\right )$ denotes the ground truth edge label on the $k$-th semantic category at pixel $p$ and $E_{k}\left (  p\right )$ indicates the predicted edge probability on the $k$-th semantic category at pixel $p$. $L_{e}$ measures the difference between $E$ and the semantic edge ground truth $G$.

\subsubsection{Decomposed Cross-Task Consistency Loss\label{sec:dctcl}}
$\Gamma_{\psi_{1}\psi_{2}}$ is modeled as a spatial gradient operation formulated as Equation \ref{formula:adaptive_pool}. Taking $S$ as the input of $\Gamma_{\psi_{1}\psi_{2}}$ can get another semantic edge probability map \begin{math} C \in R^{H\times W\times K} \end{math}. $C$ and $E$ should be consistent. Instead of directly penalizing the difference between $C$ and $E$, we penalize the difference between $C$ and $G$, $E$ and $G$ separately, thus indirectly forcing the alignment between $C$ and $E$. The formulation is as follows:
\begin{equation}
    L_{c}^{d}=\sum_{k} \sum_{p} W_{k}\left ( p \right ) \left(\mid C_{k}\left ( p \right ) -G_{k}\left ( p \right ) \mid+\mid E_{k}\left ( p \right ) -G_{k}\left ( p \right ) \mid\right).
\label{formula:ctcl}
\end{equation}
We call $L_{c}^{d}$ the decomposed cross-task consistency loss, in which
\begin{equation}
    W_{k}\left ( p \right ) =\left\{\begin{array}{cl}
    \beta^{k}, & G_{k}\left ( p \right ) =1 \\
    1-\beta^{k}, & G_{k}\left ( p \right ) =0
    \end{array}\right.,
\label{formula:beta}
\end{equation}
where \begin{math}\beta^{k}=\left|Y_{-}^{k}\right| /|Y^{k}|\end{math} and 
\begin{math}1-\beta^{k}=\left|Y_{+}^{k}\right| /|Y^{k}|\end{math}. $\left|Y_{+}^{k}\right|$
and $\left|Y_{-}^{k}\right|$ denote the edge and non-edge ground truth label sets of the $k$-th class semantic edge, respectively. Similar to $E_{k}\left( p \right)$,  $C_{k}\left( p \right)$ denotes another predicted edge probability on the $k$-th semantic category at pixel $p$.

The right-hand side of equation \ref{formula:ctcl} satisfies the following equation:
\begin{equation}
\begin{split}
    &\sum_{k} \sum_{p} W_{k}\left ( p \right ) \left(\mid C_{k}\left ( p \right ) -G_{k}\left ( p \right ) \mid +\mid E_{k}\left ( p \right ) -G_{k}\left ( p \right ) \mid\right)\\
    &=\sum_{k} \sum_{p} W_{k}\left ( p \right ) \mid C_{k}\left ( p \right ) -G_{k}\left ( p \right )  \mid+ \sum_{k} \sum_{p} W_{k}\left ( p \right )  \mid E_{k}\left ( p \right ) -G_{k}\left ( p \right )  \mid.
\end{split}
\label{formula:distributive_law}
\end{equation}
For simplicity, we define the following equations.
\begin{equation}
    L_{c 1}=\sum_{k} \sum_{p} W_{k}\left ( p \right ) \mid C_{k}\left ( p \right ) -G_{k}\left ( p \right ) \mid.
\label{formula:L_c1}
\end{equation}
\begin{equation}
    L_{c 2}=\sum_{k} \sum_{p} W_{k}\left ( p \right )  \mid E_{k}\left ( p \right ) -G_{k}\left ( p \right )  \mid.
\label{formula:L_c2}
\end{equation}
Equations \ref{formula:L_c1} and \ref{formula:L_c2} are variants of the $l_{1}$ norm, and we call this kind of variant the boundary-aware $l_{1}$ norm. Substituting equations \ref{formula:L_c1}, \ref{formula:L_c2} and \ref{formula:distributive_law} into \ref{formula:ctcl}, we derive:
\begin{equation}
    L_{c}^{d}=L_{c 1}+L_{c 2}.
\label{formula:Lc_separate}
\end{equation}

\subsubsection{Loss Function}
We perform a weighted sum of the above three losses to obtain the loss to predict domain $\psi_{1}$ from $\chi$ while enforcing the consistency with domain $\psi_{2}$ as:
\begin{equation}
    L=C_{s} L_{s}+C_{e} L_{e}+C_{c} L_{c}^{d},
\label{formula:L_total}
\end{equation}
in which $C_{s}$, $C_{e}$, $C_{c}$ are hyperparameters. As pointed out by \cite{vandenhende2021multi}, grid search is competitive or better compared to existing task balancing techniques in determining the weights of the loss functions. Therefore, in our experiments, $C_{s}$, $C_{e}$, $C_{c}$ are obtained by grid search. 

First, we generate grids representing various coefficient values that we wish to explore and search over during our experiments. The loss function $L_{s}$ primarily computes the loss for the majority of pixels in the image, leading to relatively larger loss values compared to the other two loss functions, which are specifically designed for focusing on object edges. However, we aim to prevent these two losses from being overshadowed due to their smaller values. To achieve this, we assign larger coefficients to the edge-related loss functions, prompting the model to pay closer attention to the edges during training. Specifically, we set $C_{s} \in \left \{1 \right \} $ and $C_{e}, C_{c} \in \left \{5, 10, 20 \right \} $. Then we try all combinations of the hyperparameter values from the defined grids. Since we have only one value for $C_{s}$ and three values for both $C_{e}$ and $C_{c}$
 , we have a total of 1 x 3 x 3 = 9 combinations to try.

Substituting equation \ref{formula:Lc_separate} into \ref{formula:L_total}, we obtain:
\begin{equation}
    L =C_{s} L_{s}+C_{e} L_{e}+C_{c} \left(L_{c 1}+L_{c 2}\right),
\label{formula:Ltotal_Lc_separate}
\end{equation}
which is equivalent to the following equation:
\begin{equation}
    L =\left(C_{s} L_{s}+C_{c} L_{c 1}\right)+\left(C_{e} L_{e}+C_{c} L_{c 2}\right),
\label{formula:L_total_separate}
\end{equation}
where the first term is pertinent to network $\Gamma_{\chi\psi_{1}}$, while the second term is pertinent to network $\Gamma_{\chi\psi_{2}}$. These two terms are independent and can be dealt with in parallel for task-specific layers in network $\Gamma_{\chi\psi_{1}}$ and $\Gamma_{\chi\psi_{2}}$, which is exactly the original intention of our definition of $L_{c}^{d}$ as two independent parts.

\section{Experiments}

We first conducted experiments on Cityscapes \cite{cordts2016cityscapes} which is a popular computer vision dataset for semantic urban scene understanding. It contains 5000 annotated images with fine annotations collected from 50 cities in different seasons. The images were divided into sets numbered 2,975, 500, and 1,525 for training, validation, and testing. Conventionally, only 19 categories are used to assess the accuracy of category segmentation. Although it also provides coarsely annotated images, we only use finely annotated images. In addition, experiments on FloodNet \cite{rahnemoonfar2021floodnet} were also performed to further confirm the generalization and application value of our method. Code and models are available at: https://github.com/nailperry-zd/PaddleSeg-TriangleNet.

\subsection{Experiments on Cityscapes}

\textbf{Baseline:} We append the PPM \cite{zhao2017pyramid} and FPN \cite{lin2017feature} decoder to naive FCN as the baseline, where ResNet-18 \cite{he2016deep} serves as the backbone, that is, training the semantic segmentation branch shown in \cref{Figure:pipeline} independently.

\textbf{Implementation details:} We use the 2.3.0 version of the PaddlePaddle \cite{ma2019paddlepaddle} framework to carry out the following experiments. The hardware platform adopts a single V100 GPU with a video memory of 32G. All networks with ResNet-18 \cite{he2016deep} as the backbone share some settings, where stochastic gradient descent (SGD) with a batch size of 4 is used as the optimizer, with a momentum of 0.9 and weight decay of 5e-4. All these ResNet-18 variants are trained for 300K batch iterations with an initial learning rate of 0.01. Data augmentation contains normalization, random distortion, random horizontal flip, random resizing with a scale range of [0.5, 2.0] and random cropping with a crop size of 1024 x 1024. During inference, we use the whole picture as input. In terms of loss weights, $C_{s}$, $C_{e}$ and $C_{c}$ are set to 1, 10, 20, respectively. For quantitative evaluation, mIoU is used for accuracy comparison.

\subsubsection{Comparison against state-of-the-art methods} 

We present a comprehensive comparison of our method with both real-time and non-real-time semantic segmentation algorithms in \cref{Table:stateoftheart} and \cref{Table:stateoftheart-2}, respectively.

In \cref{Table:stateoftheart}, it is important to highlight that some of the models listed have been officially integrated into PaddlePaddle and are available in the PaddleSeg \cite{liu2021paddleseg} open source library. This integration facilitates a rigorous evaluation of the inference speed for these PaddlePaddle-integrated models, as well as our model, TriangleNet. To measure the speed accurately, we utilize the PaddleInference API from the PaddleSeg \cite{liu2021paddleseg} library on an A100 GPU with 40GB memory, using the f32 accuracy parameter. However, in cases where certain models do not have a PaddlePaddle implementation or when our direct measurements are not available, we provide FPS data from the original papers or third-party sources within brackets in the table for reference. This meticulous approach ensures a comprehensive and fair comparison, allowing us to draw reliable conclusions regarding TriangleNet's performance in real-time semantic segmentation in comparison to other state-of-the-art models.

For the real-time comparison, we ensure a fair assessment by utilizing our best model based on ResNet-18 with varying inference sizes: 512 x 1024, 768 x 1536, and 1024 x 2048, represented by TriangleNet$^{1}$-0.5, TriangleNet$^{1}$-0.75, and TriangleNet$^{1}$-1.0, respectively. For the non-real-time comparison, we adopt multi-scale inference, denoted by TriangleNet$^{1}$-MS, incorporating scales of 0.75, 1.0, and 1.25. 

As shown in \cref{Table:stateoftheart}, at a resolution of 512 x 1024, our model not only surpasses ESPNetV2 in both speed and accuracy but also outperforms STDC1-Seg50 and PP-LiteSeg-T1 in accuracy, despite achieving approximately 60\% and 70\% of their respective speeds. Similarly, at 768 x 1536 resolution, while our model maintains 85\% of BiSeNetV1-L's speed, it exhibits a 1.4\% increase in accuracy compared to it. Additionally, our model's speed, at around 50\% of STDC1-Seg75 and PP-LiteSeg-T2, is compensated by its approximately 1\% higher accuracy over them. Under a resolution of 1024x2048, our model exhibits significantly higher accuracy than ICNet, SwiftNet, and FasterSeg. 

Our method demonstrates a remarkable speed/accuracy trade-off across various resolutions when compared to real-time counterparts. Notably, our model achieves impressive accuracy without compromising on speed, enabling real-time inference even at full resolution. Notably, in \cref{Table:stateoftheart-2}, our model demonstrates competitive performance even compared to non-real-time models based on ResNet-101 \cite{he2016deep}, achieving similar mIoU scores while utilizing only one-fifth of the parameters. This highlights the efficiency and effectiveness of our approach across diverse scenarios.

\begin{table}[ht!]
\centering
\scriptsize
\begin{tabular}{c c c c c c}
    \toprule
    
    \multirow{2}{*}{Model} & \multirow{2}{*}{Backbone}& \multirow{2}{*}{mIoU} & \multicolumn{3}{c}{FPS}   \\
        \cline{4-6}
          & & & 1024x2048 & 768x1536 & 512x1024\\
    \midrule
    ICNet \cite{zhao2018icnet} & PSPNet50 & 69.5 & -(30.3) &  &  \\
    ESPNetV2 \cite{mehta2019espnetv2} & - & 66.2 &  &  & 126.5(114.7+) \\
    SwiftNet \cite{wang2021swiftnet} & ResNet18 & 75.5 & -(39.9) &&\\
    BiSeNetV1 \cite{yu2018bisenet} & Xception39 & 68.4 &&-(105.8)&\\
    BiSeNetV1-L \cite{yu2018bisenet} & ResNet18 & 74.7 &  & 83.9(65.5) & \\
    BiSeNetV2 \cite{yu2021bisenet} &- & 72.6 &&&-(156) \\
    BiSeNetV2-L \cite{yu2021bisenet} & -  & 75.3 &&&-(47.3) \\
    FasterSeg \cite{chen2019fasterseg} & -  & 71.5 & -(163.9) &&\\
    STDC1-Seg50 \cite{fan2021rethinking} & STDC1 & 71.9 &&&262.1(250.4) \\
    STDC2-Seg50 \cite{fan2021rethinking} & STDC2 & 73.4 &&&207.4(188.6) \\
    STDC1-Seg75 \cite{fan2021rethinking} & STDC1 & 75.3 &&152.7(126.7)& \\
    STDC2-Seg75 \cite{fan2021rethinking} & STDC2 & 76.8 &&131.5(97.0)& \\
    PP-LiteSeg-T1 \cite{peng2022pp} &STDC1  & 72.0 &&&219.4(273.6) \\
    PP-LiteSeg-B1 \cite{peng2022pp} &STDC2 & 73.9 &&&184.3(195.3) \\
    PP-LiteSeg-T2 \cite{peng2022pp} &STDC1 & 74.9 &  & 141.2(143.6) & \\
    PP-LiteSeg-B2 \cite{peng2022pp} & STDC2 & 77.5 &  & 118.4(102.6) & \\
    \midrule
    TriangleNet$^{1}$-0.5 & ResNet18 & 72.7 &  & & 157.4\\
    TriangleNet$^{1}$-0.75 & ResNet18 & 76.1 &  & 71.0 & \\
    TriangleNet$^{1}$-1.0 & ResNet18 & 77.4 & 46.2 &  & \\
    \bottomrule
\end{tabular}
\caption{Accuracy comparison of our best models based on ResNet-18 against real-time models on the Cityscapes test dataset. "-" indicates that the corresponding data are not given. \textbf{FPS}: frames per second. \textbf{TriangleNet$^{1}$} is an instance of the framework shown in \cref{Figure:pipeline}. In the three columns of FPS, the values outside the brackets are measured by our team, whereas the values within the brackets are either sourced from the original papers or from third-party papers. "+" denotes the value is sourced from \cite{holder2022efficient}.}
\label{Table:stateoftheart}
\end{table}

\begin{table}[ht!]
\centering
\begin{tabular}{c c c c c}
    \toprule
    Model & Backbone & Params(M) & mIoU(\%) \\
    \midrule
    DeepLab \cite{chen2014semantic} & ResNet-101 \cite{he2016deep} & 59 & 63.1 \\
    DepthSeg \cite{kong2018recurrent} & ResNet-101 & 58 & 78.2 \\
    PSPNet \cite{zhao2017pyramid} & ResNet-101 & 65 & 78.4\\
    \midrule
    TriangleNet$^{1}$-MS & ResNet-18 & 13 &77.8\\
    \bottomrule
\end{tabular}
\caption{Accuracy comparison of our best model based on ResNet-18 against non-real-time models on the Cityscapes test dataset.}
\label{Table:stateoftheart-2}
\end{table}

\subsubsection{Ablation Studies\label{sec:ablation}}
Our approach involves several elements compared to the baseline. Each element may contribute to the improvement of mIoU. To verify the necessity of each element, we performed the following ablation studies.

\textbf{Ablation study on joint framework}: From a multi-task perspective, joint training benefits from higher task correlation. To explore this idea, we conducted experiments combining different tasks. In \cref{Table:joint}, the second row demonstrates joint training of Baseline with HED \cite{xie2015holistically}, a classic binary edge detection model, resulting in a slight improvement in mIoU on the Cityscapes test set. Subsequently, we replaced HED with DFF \cite{hu2019dynamic}, a superior semantic edge detection model, in the third row. This change led to a 0.59\% improvement against the Baseline in mIoU on the Cityscapes test set. The results suggest a stronger correlation between semantic segmentation and semantic edge detection tasks. This correlation arises from the accurate extraction of semantic edges under the constraints of semantic segmentation, as semantic segmentation can suppress non-edge pixels, and in turn, relies on semantic edges to distinguish between objects and background. The two tasks mutually complement each other, enhancing the overall performance of the model.

\begin{figure}[ht]
  \centering
  \includegraphics[width=0.8\linewidth]{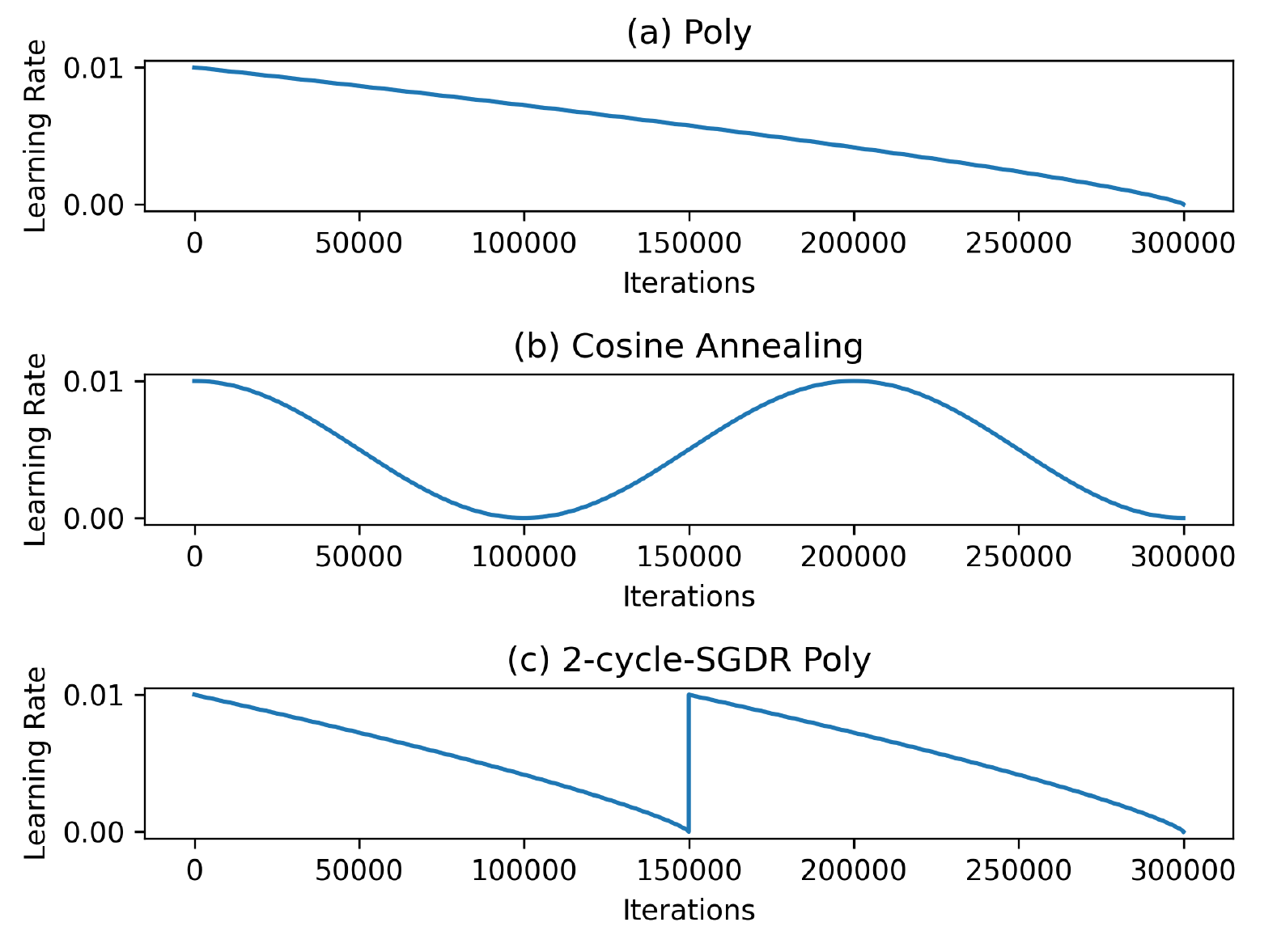}
  \caption{Visualization of different learning rate policies.}
  
  \label{Figure:lr3}
\end{figure}

During this process, we adopted the Poly learning rate policy, which is widely used and proven effective, as depicted in \cref{Figure:lr3} (a).

\begin{table}[ht!]
    \centering
    \begin{tabular}{c c c c}
    \toprule
    \multirow{2}{*}{Model} & \multirow{2}{*}{LR Policy} &  \multicolumn{2}{c}{mIoU(\%)} \\
        \cline{3-4}
         & & val & test\\
        \midrule
        Baseline & Poly & 76.77 & 74.48 \\
        Baseline+HED \cite{xie2015holistically} & Poly & 77.33  & 74.66 \\
        Baseline+DFF \cite{hu2019dynamic} \ & Poly & \textbf{78.13} & \textbf{75.07} \\
        \bottomrule
    \end{tabular}
    \caption{Ablation study on joint framework.}

\label{Table:joint}
\end{table}

\textbf{Ablation study on $L_{c}^{d}$}: We introduced the $L_{c}^{d}$ during the last 50\% iterations to verify its effect. This idea draws on \cite{wang2022active}, where a loss called ABL is added at the last 20\% epochs, since the gradient of ABL is not useful when the semantic edges output by the network are far from the semantic edge ground truth at the beginning of the training, much similar to our case.

\begin{table}[ht!]
    \centering
    \scalebox{0.8}{
    \begin{tabular}{c c c c c}
    \toprule
    \multirow{2}{*}{Model} & \multirow{2}{*}{LR Policy} & \multirow{2}{*}{L$_{c}^{d}$ involved} & \multicolumn{2}{c}{mIoU(\%)} \\
        \cline{4-5}
          & & & val & test\\
        \midrule
        Baseline+DFF & Poly & $\times $ &78.13 & 75.07 \\
        Baseline+DFF & 2-cycl-SGDR Poly & $\times $ &78.41(0.28 $\uparrow $) & 75.53(0.46 $\uparrow $) \\
        TriangleNet$^{1}$ &2-cycl-SGDR Poly& $\surd $  & \textbf{78.96(0.55 $\uparrow $)} & \textbf{77.36(1.83 $\uparrow $)} \\
        TriangleNet$^{1}$ &Cosine Annealing& $\surd $  & 78.65 & 76.93 \\
        \bottomrule
    \end{tabular}
    }
    \caption{Ablation on L$_{c}^{d}$. All these models are trained for 300K iterations. "$\times$" means that L$_{c}^{d}$ is not involved in all iterations. \textbf{TriangleNet$^{1}$} is an instance of the framework shown in \cref{Figure:pipeline}. In this situation, we get all the results by single-scale inference.}

\label{Table:lc}
\end{table}

During this process, we employed a custom learning rate policy named "2-cycle-SGDR Poly," which can be seen as a variant of the Cosine Annealing policy \cite{loshchilov2016sgdr}. The visualizations of the Cosine Annealing and 2-cycle-SGDR Poly policies are shown in \cref{Figure:lr3} (b) and \cref{Figure:lr3} (c), respectively. In the 2-cycle-SGDR Poly policy, the learning rate periodically increases, a process referred to as "restarts" in SGDR \cite{loshchilov2016sgdr}. The underlying idea is to encourage the model to traverse from one local minimum to another, particularly if it is trapped in a steep trough. 

After comparing the first and second rows in \cref{Table:lc}, we observed that employing the 2-cycle-SGDR Poly policy alone resulted in a 0.46\% increase in mIoU on the Cityscapes test set. Subsequently, with the introduction of $L_{c}^{d}$ in the third row, the mIoU exhibited a consistent growth trend on both the validation and test sets, with a more significant improvement observed on the test set. The inclusion of $L_{c}^{d}$ further boosted the mIoU by 1.83\% on the test set, indicating its positive impact on the overall performance of our model.

Furthermore, upon comparing the third and fourth rows in \cref{Table:lc}, we found that both Cosine Annealing and 2-cycle-SGDR Poly policies can improve the model to some extent, confirming the effectiveness of the "restarts" in SGDR. Notably, 2-cycle-SGDR Poly is more suitable for our model. Therefore, for all subsequent TriangleNet variants, we adopted the 2-cycle-SGDR Poly learning rate schedule to ensure consistent and superior optimization of our model.

\textbf{Ablation study on different semantic edge detection strategies:} In our exploration of state-of-the-art strategies for semantic edge detection, we conducted a comparison between DFF \cite{hu2019dynamic} and CASENet \cite{yu2017casenet}. The two rows in \cref{Table:sed} demonstrate that both DFF and CASENet, when employed as semantic edge detection branches, lead to improved segmentation accuracy. This finding highlights the positive role of injecting semantic edges into the semantic segmentation process. Notably, as mentioned in \cite{hu2019dynamic}, DFF surpasses CASENet in standalone semantic edge extraction. Moreover, even after integrating with semantic segmentation, DFF continues to deliver superior accuracy improvements, reaffirming its effectiveness in enhancing the overall performance of the model.

\begin{table}[ht!]
    \centering
    \begin{tabular}{c c c c}
    \toprule
    \multirow{2}{*}{Model} & \multirow{2}{*}{SED strategy} & \multicolumn{2}{c}{mIoU(\%)} \\
        \cline{3-4}
          & & val & test\\
        \midrule
        TriangleNet$^{1}$ & DFF \cite{hu2019dynamic}  &\textbf{78.96} & \textbf{77.36} \\
        TriangleNet$^{2}$ & CASENet \cite{yu2017casenet}  &78.87 & 77.11 \\

        \bottomrule
    \end{tabular}
    \caption{Experiments on different semantic edge detection (SED) strategies. \textbf{TriangleNet$^{2}$} is the same as TriangleNet$^{1}$ except that TriangleNet$^{2}$
   uses CASENet \cite{yu2017casenet} as the semantic edge detection. In this situation, we get all the results by single-scale inference.}

\label{Table:sed}
\end{table}

\textbf{Ablation study on different semantic segmentation models:} Our evaluation extends to various semantic segmentation models, and the results presented in \cref{Table:ssmodels} reveal that the integration of U-Net \cite{ronneberger2015u} and Baseline into our framework for joint training yields remarkable improvements in semantic segmentation accuracy, showcasing the effectiveness and benefits of our approach in enhancing semantic segmentation performance.

\begin{table}[ht!]
\centering
\begin{tabular}{c c}
    \toprule
    Model& mIoU(\%) on val \\
    \midrule
    U-Net \cite{ronneberger2015u} & 66.34  \\
    TriangleNet$^{3}$(SEM:U-Net)  & 69.12(2.78 $\uparrow $) \\
    Baseline & 76.77\\
    TriangleNet$^{1}$(SEM:Baseline)& 78.96(2.19 $\uparrow $)\\
    \bottomrule
\end{tabular}
\caption{Experiments on different semantic segmentation (SEM) strategies. \textbf{TriangleNet$^{3}$} is the same as TriangleNet$^{1}$ except that TriangleNet$^{3}$
   uses U-Net \cite{ronneberger2015u} as the semantic segmentation branch.}
\label{Table:ssmodels}
\end{table}

\subsubsection{Analyses}
The results from the ablation studies demonstrate that the semantic edge detection task exhibits a stronger correlation with the semantic segmentation task compared to the binary edge detection task. When we jointly train both tasks, we observe a 0.59\% improvement in mIoU on the Cityscapes test set. Additionally, the adoption of the 2-cycle-SGDR Poly learning rate policy leads to a slight yet meaningful improvement of 0.46\%. 

However, despite the benefits of multi-task learning, the implicit learning of cross-task consistency in multi-task networks is limited. To address this limitation, we introduced an additional restriction called decoupled cross-task consistency loss, denoted as $L_{c}^{d}$, to explicitly enhance cross-task consistency between semantic edge detection and semantic segmentation.

The explicit enhancement of consistency through this restriction resulted in a further significant improvement of 1.83\% in mIoU on the Cityscapes test set. These findings underscore the importance of incorporating explicit consistency constraints during the learning process to achieve enhanced performance in multi-task computer vision systems.

Upon analyzing the IoU of each category, as depicted in \cref{Table:details}, we observe significant improvements for several categories, with some experiencing increases of more than 3\%. Notably, the largest improvement amounts to nearly 18\% in IoU score. This further validates the significance of incorporating edge information through our edge prior augmentation approach, particularly for categories such as "truck," "bus," and "train," where distinct edges and boundaries are prevalent.

\begin{figure}[ht]
  \centering
  \includegraphics[width=0.95\linewidth]{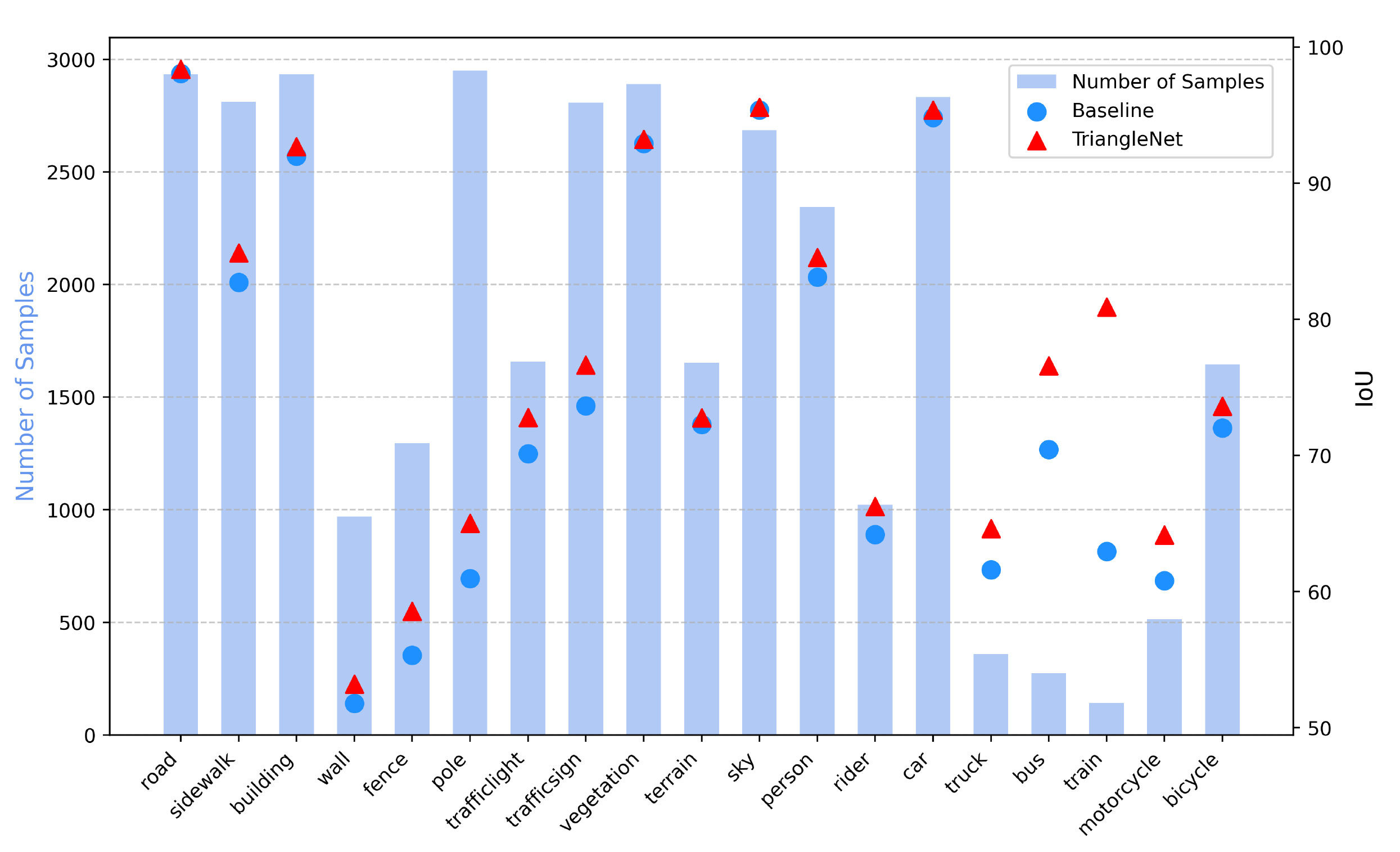}
  \caption{Category-level analysis of samples and intersection over union (IoU) on Cityscapes test dataset.}
  
  \label{Figure:analysis_on_cityscapes}
\end{figure}

Furthermore, we investigate the relationship between the number of sample images per category and their respective IoU scores, visualized in \cref{Figure:analysis_on_cityscapes}. The analysis reveals that categories with higher IoUs tend to have more samples, while those with lower IoUs have fewer samples, aligning with our expectations. Notably, categories such as "truck," "bus," and "train," despite having smaller sample sizes, exhibit remarkable improvements in IoU. This suggests that TriangleNet can effectively generalize from limited samples, resulting in enhanced performance in challenging categories.

However, certain categories, such as "wall," "fence," "pole," and "trafficsign," possess abundant samples but fail to achieve the expected higher IoU values. The underperformance of these categories may be attributed to factors such as complex semantic patterns or limitations in the model architecture to accurately capture their unique features. Further investigation is warranted to identify the specific reasons behind these discrepancies and to devise strategies to enhance the segmentation performance for these categories.

To summarize, TriangleNet demonstrates a remarkable 2.88\% improvement in mean Intersection over Union (mIoU) on the Cityscapes test dataset compared to the Baseline. Our model outperforms the Baseline in all categories, particularly in scenarios with distinct edges and boundaries, substantiating the efficacy of multi-task learning and explicit cross-task consistency enhancement.

\begin{table}[ht!]
\centering
\scalebox{0.38}{
\begin{tabular}{c c c c c c c c c c c c c c c c c c c c c}
    \toprule
    Model & road& sidewalk& building& wall& fence& pole& trafficlight& trafficsign& vegetation& terrain& sky& person& rider& car& truck& bus& train& motorcycle& bicycle & mIoU\\
    \midrule
    Baseline & 98.05& 82.70& 92.99& 51.81& 55.34& 60.97& 70.13& 73.64& 92.93& 72.29& 95.37& 83.09& 64.19& 94.84& 61.62& 70.45& 62.96& 60.81& 72.03& 74.48\\
    Baseline+DFF & 98.32& 84.39& 92.17& 48.92& 57.15& 62.34& 71.34& 75.48& 93.12& 73.10& 95.49& 83.41& 63.85& 95.16& 63.11& 69.83& 63.91& 62.40& 72.80& 75.07\\
    TriangleNet$^{1}$ & 98.37& 84.87& 92.68& 53.22& 58.56& 65.01& 72.79& 76.63& 93.22& 72.77& 95.59& 84.55& 66.26& 95.37& 64.63& 76.59& 80.91& 64.16& 73.63& 77.36\\
    \midrule
    & 0.32& 2.17& 0.69& 1.41& \textbf{3.23}& \textbf{4.04}& 2.66& \textbf{2.99}& 0.29& 0.48& 0.22& 1.46& 2.07& 0.54& \textbf{3.01}& \textbf{6.15}& \textbf{17.94}& \textbf{3.35}& 1.60& 2.87\\
    \bottomrule
    
\end{tabular}
}
\caption{Intersection over union (IoU) growth per category on the Cityscapes test set. }
\label{Table:details}
\end{table}

\subsubsection{Visualization}
\begin{figure}[ht]
  \centering
  \includegraphics[width=0.95\linewidth]{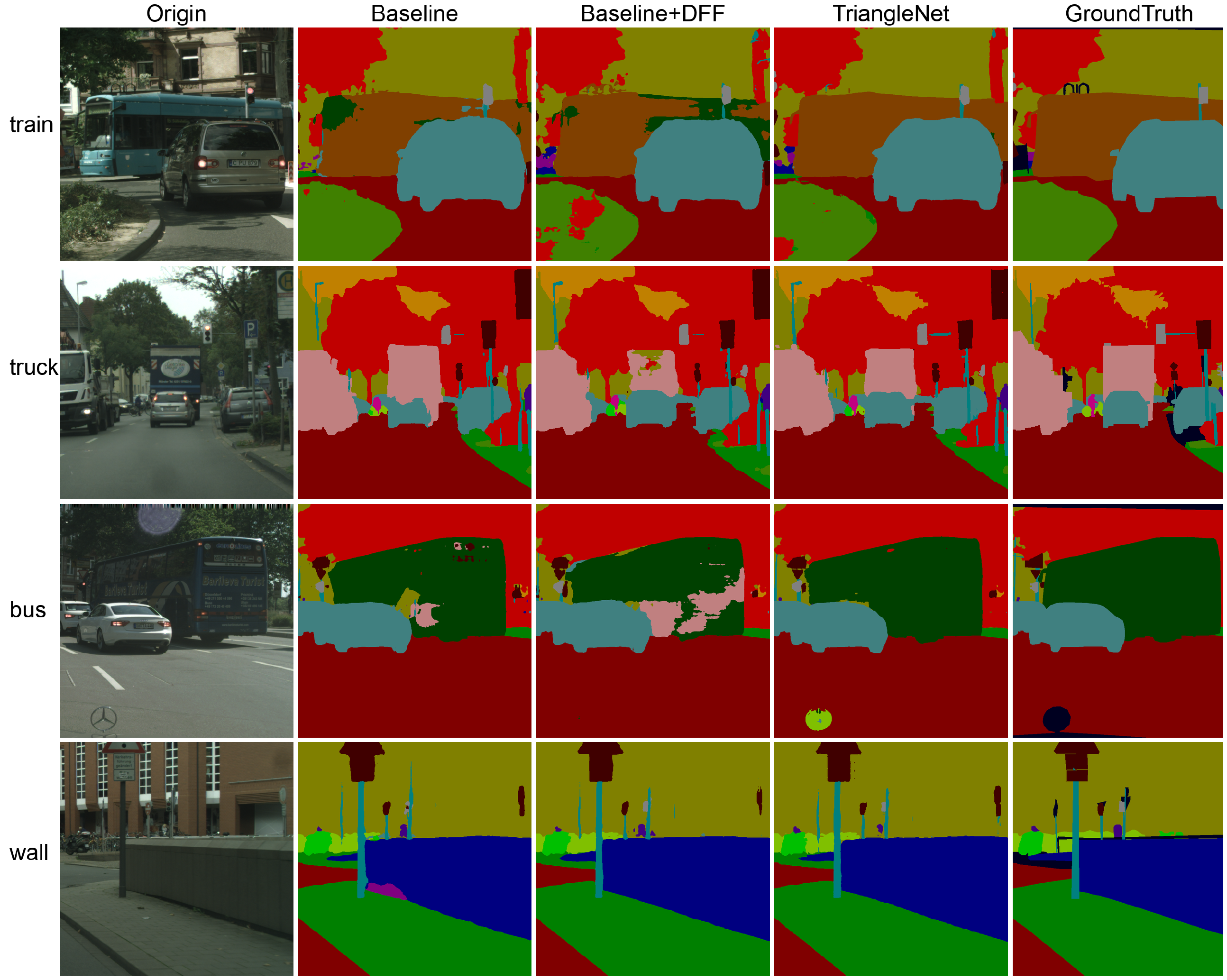}
  \caption{Visualization on Cityscapes.}
  
  \label{Figure:visual}
\end{figure}

We performed a qualitative comparison by visualizing segmentation maps of various categories. In \cref{Figure:visual}, we present the segmentation maps generated by different models for this purpose. Notably, in the third row, TriangleNet demonstrates its ability to accurately locate pixels along two objects by effectively utilizing edge priors or shapes of objects, which the Baseline fails to achieve. Additionally, in the first, third, and fourth rows, TriangleNet leverages the priors obtained from semantic edge detection to perceive trains, buses, and walls as coherent entities, while the Baseline tends to split them into different categories. This highlights TriangleNet's capacity to benefit from edge information and enhance its understanding of complex object structures, resulting in improved semantic segmentation performance.

\subsection{Experiments on FloodNet}
We also performed experiments on FloodNet \cite{rahnemoonfar2021floodnet} which is an Unmanned Aerial Vehicle (UAV) dataset to assess the damage from natural disasters to further prove the compatibility of our method. The dataset contains 2343 images in total, which were divided into sets numbered 1,445, 450, and 448 for training, validation, and testing. This dataset contains 10 classes, and the index and specific meaning of each class are given by \cref{Table:floodnet}(a). 

\textbf{Implementation details:} We verified the performance of the Baseline and TriangleNet aforementioned on the FloodNet dataset. Different from experiments on Cityscapes, some hyper-parameters need to be changed. Specifically, the batch size of SGD is set to 16, and all these ResNet-18 \cite{he2016deep} variants are trained for 20K iterations with 4 V100 GPUs. In terms of loss weights, $C_{s}$, $C_{e}$ and $C_{c}$ are set to 1, 2, 4, respectively.

We analysed prominantly improved categories on FloodNet. As shown in \cref{Table:floodnet}(b), TriangleNet still achieves better performance on all categories against the Baseline. It's worth noting that although DeepLabV3+ \cite{chen2018encoder} is based on ResNet-101 \cite{he2016deep}, its acruacy on FloodNet is not as good as our model based on ResNet-18 \cite{he2016deep}.

\begin{table}[ht]
    \begin{subtable}[t]{1\linewidth}
    \centering
    \scalebox{0.6}{
        \begin{tabular}{c c c}
            \toprule
            Index & Class & Images\\
            \midrule
            0 & Background & -\\
            1 & Building-flooded & 275\\
            2 & Building-non-flooded & 1272\\
            3 & Road-flooded & 335\\
            4 & Road-non-flooded & 1725\\
            5 & Water & 1262\\
            6 & Tree & 2507 \\
            7 & Vehicle & 1105\\
            8 & Pool & 676\\
            9 & Grass &-\\
            \bottomrule
        \end{tabular}
        }
        \caption{10 classes of FloodNet}
    \end{subtable}

    \begin{subtable}[t]{1\linewidth}
        \centering
        \scalebox{0.6}{
        \begin{tabular}{c c c c c c c c c c c c}
            \toprule
            Method & Backbone & 1 & 2 & 3 & 4 & 5 & 6 & 7 & 8 & 9 & mIoU \\
            \midrule
            DeepLabV3+ \cite{chen2018encoder} & ResNet-101 & 32.7 & 72.8 & 52 & 70.2 & \textbf{75.2} & 77.0 & 42.5 & 47.1 & 84.3 & 61.53 \\
            Baseline & ResNet-18 & 72.58 & 73.86 & 53.68 & 80.05 & 69.36 & 79.06 & 57.8 & 57.06 & 87.36 & 65.64 \\
            TriangleNet$^{1}$ & ResNet-18& \textbf{78.26} &\textbf{75.38} &\textbf{56.44} &\textbf{82.45} &74.57 &\textbf{82.86} &\textbf{61.33} &\textbf{61.73} &\textbf{89.57} & \textbf{70.97}\\
            \bottomrule
        \end{tabular}
        }
        \caption{Per-class results on the FloodNet testing set. We obtain the results of the last two rows by single-scale inference.}
    \end{subtable}

    \caption{Experiments on the FloodNet testing set.}
    \label{Table:floodnet}
\end{table}

\begin{figure}[ht]
  \centering
  \includegraphics[width=0.95\linewidth]{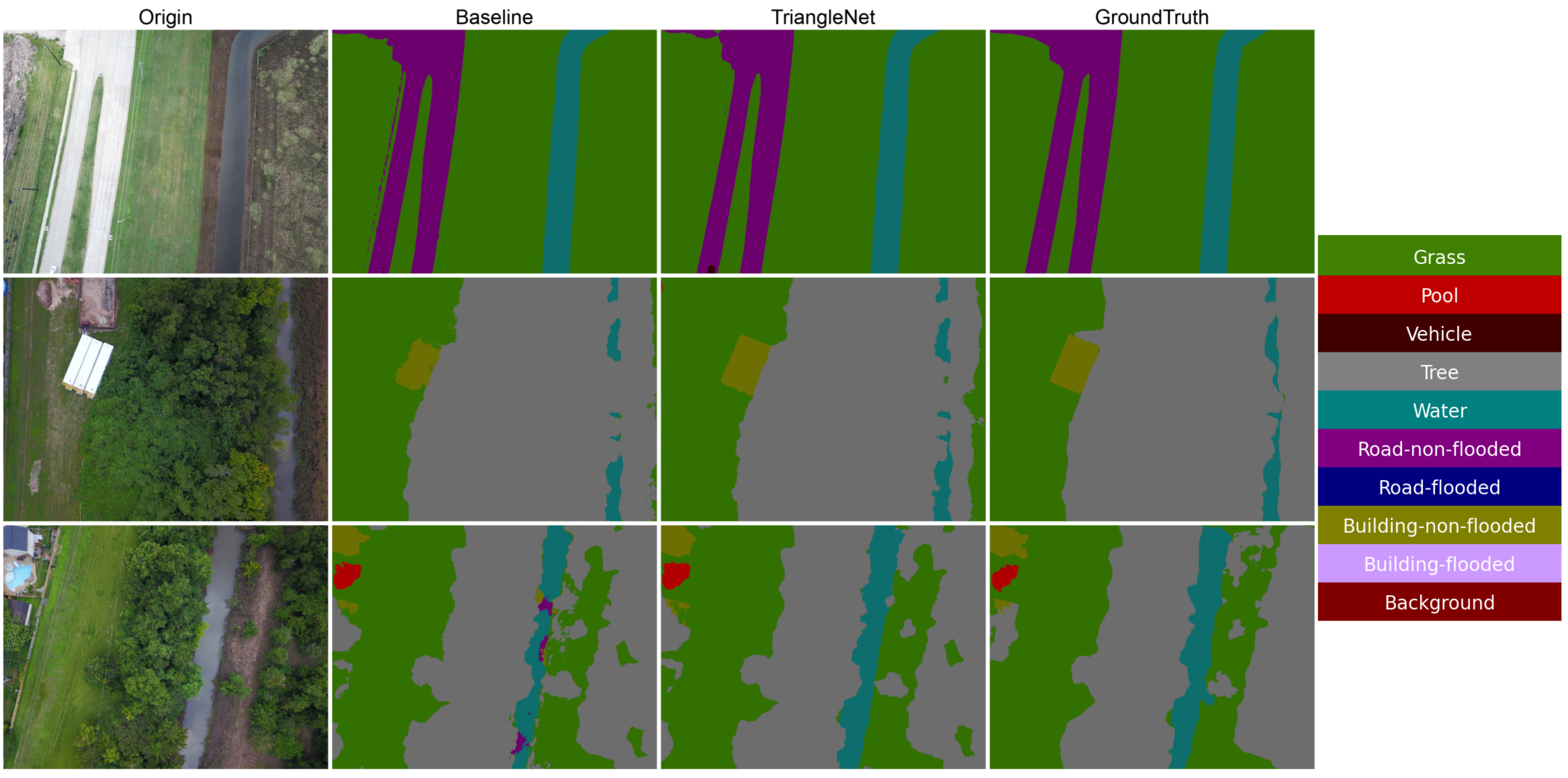}
  \caption{Visualization on FloodNet.}
  
  \label{Figure:samples_floodnet}
\end{figure}

The improvement can obviously be visualized in \cref{Figure:samples_floodnet}. In the first row, the non-flooded building is well segmented in TriangleNet. In the second row, the water is perceived as a whole, while Baseline divides it into parts. In the third row, TriangleNet is able to segment the sharp edge of the non-flooded road, while Baseline fails.

\section{Conclusion}
In this paper, we presented TriangleNet, an innovative model that utilizes a decoupled architecture for joint training of multiple tasks without the need for fusion modules during inference. This design allows our model to reap the benefits of multitasking during training while avoiding any additional inference time, making TriangleNet a practical and efficient solution for real-time semantic segmentation. By employing multi-task learning and explicit cross-task consistency enhancement, TriangleNet consistently achieves improvements on both the Cityscapes and FloodNet datasets, showcasing its robust generalization capabilities in various environmental conditions.

In summary, TriangleNet showcases its potential in semantic segmentation by effectively leveraging edge priors and incorporating explicit cross-task consistency. This unique combination not only enhances accuracy but also enables real-time inference, making it well-suited for various real-world applications. Further research exploring more detailed explicit constraints may lead to even greater performance improvements. The achievements of TriangleNet in the context of real-time semantic segmentation paves the way for future advancements in efficient and accurate computer vision systems.

\bibliographystyle{unsrt}
\bibliography{main}

\begin{thebibliography}{10}

\bibitem{long2015fully}
Jonathan Long, Evan Shelhamer, and Trevor Darrell.
\newblock Fully convolutional networks for semantic segmentation.
\newblock In {\em Proceedings of the IEEE Conference on Computer Vision and
  Pattern Recognition (CVPR)}, pages 3431--3440, 2015.

\bibitem{ronneberger2015u}
Olaf Ronneberger, Philipp Fischer, and Thomas Brox.
\newblock U-net: Convolutional networks for biomedical image segmentation.
\newblock In {\em International Conference on Medical image computing and
  computer-assisted intervention}, pages 234--241. Springer, 2015.

\bibitem{wu2019fastfcn}
Huikai Wu, Junge Zhang, Kaiqi Huang, Kongming Liang, and Yizhou Yu.
\newblock Fastfcn: Rethinking dilated convolution in the backbone for semantic
  segmentation.
\newblock {\em arXiv preprint arXiv:1903.11816}, 2019.

\bibitem{takikawa2019gated}
Towaki Takikawa, David Acuna, Varun Jampani, and Sanja Fidler.
\newblock Gated-scnn: Gated shape cnns for semantic segmentation.
\newblock In {\em Proceedings of the IEEE international conference on computer
  vision (ICCV)}, pages 5229--5238, 2019.

\bibitem{chen2014semantic}
Liang-Chieh Chen, George Papandreou, Iasonas Kokkinos, Kevin Murphy, and Alan~L
  Yuille.
\newblock Semantic image segmentation with deep convolutional nets and fully
  connected crfs.
\newblock {\em arXiv preprint arXiv:1412.7062}, 2014.

\bibitem{chen2017deeplab}
Liang-Chieh Chen, George Papandreou, Iasonas Kokkinos, Kevin Murphy, and Alan~L
  Yuille.
\newblock Deeplab: Semantic image segmentation with deep convolutional nets,
  atrous convolution, and fully connected crfs.
\newblock {\em IEEE transactions on pattern analysis and machine intelligence},
  40(4):834--848, 2017.

\bibitem{chen2017rethinking}
Liang-Chieh Chen, George Papandreou, Florian Schroff, and Hartwig Adam.
\newblock Rethinking atrous convolution for semantic image segmentation.
\newblock {\em arXiv preprint arXiv:1706.05587}, 2017.

\bibitem{he2017mask}
Kaiming He, Georgia Gkioxari, Piotr Doll{\'a}r, and Ross Girshick.
\newblock Mask r-cnn.
\newblock In {\em Proceedings of the IEEE international conference on computer
  vision (ICCV)}, pages 2961--2969, 2017.

\bibitem{ulku2022survey}
Irem Ulku and Erdem Akag{\"u}nd{\"u}z.
\newblock A survey on deep learning-based architectures for semantic
  segmentation on 2d images.
\newblock {\em Applied Artificial Intelligence}, pages 1--45, 2022.

\bibitem{ruder2017overview}
Sebastian Ruder.
\newblock An overview of multi-task learning in deep neural networks.
\newblock {\em arXiv preprint arXiv:1706.05098}, 2017.

\bibitem{zhen2020joint}
Mingmin Zhen, Jinglu Wang, Lei Zhou, Shiwei Li, Tianwei Shen, Jiaxiang Shang,
  Tian Fang, and Long Quan.
\newblock Joint semantic segmentation and boundary detection using iterative
  pyramid contexts.
\newblock In {\em Proceedings of the IEEE Conference on Computer Vision and
  Pattern Recognition (CVPR)}, pages 13666--13675, 2020.

\bibitem{liu2020auxiliary}
Wenrui Liu, Zongqing Lu, and He~Xu.
\newblock Auxiliary edge detection for semantic image segmentation.
\newblock In {\em Proceedings of the 2020 6th International Conference on
  Computing and Artificial Intelligence}, pages 182--187, 2020.

\bibitem{zamir2020robust}
Amir~R Zamir, Alexander Sax, Nikhil Cheerla, Rohan Suri, Zhangjie Cao, Jitendra
  Malik, and Leonidas~J Guibas.
\newblock Robust learning through cross-task consistency.
\newblock In {\em Proceedings of the IEEE Conference on Computer Vision and
  Pattern Recognition (CVPR)}, pages 11197--11206, 2020.

\bibitem{hao2020brief}
Shijie Hao, Yuan Zhou, and Yanrong Guo.
\newblock A brief survey on semantic segmentation with deep learning.
\newblock {\em Neurocomputing}, 406:302--321, 2020.

\bibitem{vandenhende2021multi}
Simon Vandenhende, Stamatios Georgoulis, Wouter Van~Gansbeke, Marc Proesmans,
  Dengxin Dai, and Luc Van~Gool.
\newblock Multi-task learning for dense prediction tasks: A survey.
\newblock {\em IEEE transactions on pattern analysis and machine intelligence},
  2021.

\bibitem{minaee2021image}
Shervin Minaee, Yuri~Y Boykov, Fatih Porikli, Antonio~J Plaza, Nasser
  Kehtarnavaz, and Demetri Terzopoulos.
\newblock Image segmentation using deep learning: A survey.
\newblock {\em IEEE transactions on pattern analysis and machine intelligence},
  2021.

\bibitem{yu2017dilated}
Fisher Yu, Vladlen Koltun, and Thomas Funkhouser.
\newblock Dilated residual networks.
\newblock In {\em Proceedings of the IEEE Conference on Computer Vision and
  Pattern Recognition (CVPR)}, pages 472--480, 2017.

\bibitem{sun2019deep}
Ke~Sun, Bin Xiao, Dong Liu, and Jingdong Wang.
\newblock Deep high-resolution representation learning for human pose
  estimation.
\newblock In {\em Proceedings of the IEEE Conference on Computer Vision and
  Pattern Recognition (CVPR)}, pages 5693--5703, 2019.

\bibitem{pinheiro2014recurrent}
Pedro Pinheiro and Ronan Collobert.
\newblock Recurrent convolutional neural networks for scene labeling.
\newblock In {\em International conference on machine learning}, pages 82--90.
  PMLR, 2014.

\bibitem{byeon2015scene}
Wonmin Byeon, Thomas~M Breuel, Federico Raue, and Marcus Liwicki.
\newblock Scene labeling with lstm recurrent neural networks.
\newblock In {\em Proceedings of the IEEE Conference on Computer Vision and
  Pattern Recognition (CVPR)}, pages 3547--3555, 2015.

\bibitem{zhao2017pyramid}
Hengshuang Zhao, Jianping Shi, Xiaojuan Qi, Xiaogang Wang, and Jiaya Jia.
\newblock Pyramid scene parsing network.
\newblock In {\em Proceedings of the IEEE Conference on Computer Vision and
  Pattern Recognition (CVPR)}, pages 2881--2890, 2017.

\bibitem{zhang2019dual}
Li~Zhang, Xiangtai Li, Anurag Arnab, Kuiyuan Yang, Yunhai Tong, and Philip~HS
  Torr.
\newblock Dual graph convolutional network for semantic segmentation.
\newblock {\em arXiv preprint arXiv:1909.06121}, 2019.

\bibitem{zheng2015conditional}
Shuai Zheng, Sadeep Jayasumana, Bernardino Romera-Paredes, Vibhav Vineet,
  Zhizhong Su, Dalong Du, Chang Huang, and Philip~HS Torr.
\newblock Conditional random fields as recurrent neural networks.
\newblock In {\em Proceedings of the IEEE international conference on computer
  vision (ICCV)}, pages 1529--1537, 2015.

\bibitem{wang2018non}
Xiaolong Wang, Ross Girshick, Abhinav Gupta, and Kaiming He.
\newblock Non-local neural networks.
\newblock In {\em Proceedings of the IEEE Conference on Computer Vision and
  Pattern Recognition (CVPR)}, pages 7794--7803, 2018.

\bibitem{fu2019dual}
Jun Fu, Jing Liu, Haijie Tian, Yong Li, Yongjun Bao, Zhiwei Fang, and Hanqing
  Lu.
\newblock Dual attention network for scene segmentation.
\newblock In {\em Proceedings of the IEEE Conference on Computer Vision and
  Pattern Recognition (CVPR)}, pages 3146--3154, 2019.

\bibitem{cheng2022mifnet}
Jieren Cheng, Xin Peng, Xiangyan Tang, Wenxuan Tu, and Wenhang Xu.
\newblock Mifnet: A lightweight multiscale information fusion network.
\newblock {\em International Journal of Intelligent Systems}, 37(9):5617--5642,
  2022.

\bibitem{zhang2018joint}
Zhenyu Zhang, Zhen Cui, Chunyan Xu, Zequn Jie, Xiang Li, and Jian Yang.
\newblock Joint task-recursive learning for semantic segmentation and depth
  estimation.
\newblock In {\em Proceedings of the European Conference on Computer Vision
  (ECCV)}, pages 235--251, 2018.

\bibitem{mousavian2016joint}
Arsalan Mousavian, Hamed Pirsiavash, and Jana Ko{\v{s}}eck{\'a}.
\newblock Joint semantic segmentation and depth estimation with deep
  convolutional networks.
\newblock In {\em 2016 Fourth International Conference on 3D Vision (3DV)},
  pages 611--619. IEEE, 2016.

\bibitem{nekrasov2019real}
Vladimir Nekrasov, Thanuja Dharmasiri, Andrew Spek, Tom Drummond, Chunhua Shen,
  and Ian Reid.
\newblock Real-time joint semantic segmentation and depth estimation using
  asymmetric annotations.
\newblock In {\em 2019 International Conference on Robotics and Automation
  (ICRA)}, pages 7101--7107. IEEE, 2019.

\bibitem{he2021sosd}
Lei He, Jiwen Lu, Guanghui Wang, Shiyu Song, and Jie Zhou.
\newblock Sosd-net: Joint semantic object segmentation and depth estimation
  from monocular images.
\newblock {\em Neurocomputing}, 440:251--263, 2021.

\bibitem{de2018panoptic}
Daan De~Geus, Panagiotis Meletis, and Gijs Dubbelman.
\newblock Panoptic segmentation with a joint semantic and instance segmentation
  network.
\newblock {\em arXiv preprint arXiv:1809.02110}, 2018.

\bibitem{zhao2020jsnet}
Lin Zhao and Wenbing Tao.
\newblock Jsnet: Joint instance and semantic segmentation of 3d point clouds.
\newblock In {\em Proceedings of the AAAI Conference on Artificial
  Intelligence}, volume~34, pages 12951--12958, 2020.

\bibitem{ding2019boundary}
Henghui Ding, Xudong Jiang, Ai~Qun Liu, Nadia~Magnenat Thalmann, and Gang Wang.
\newblock Boundary-aware feature propagation for scene segmentation.
\newblock In {\em Proceedings of the IEEE international conference on computer
  vision (ICCV)}, pages 6819--6829, 2019.

\bibitem{chen2016semantic}
Liang-Chieh Chen, Jonathan~T Barron, George Papandreou, Kevin Murphy, and
  Alan~L Yuille.
\newblock Semantic image segmentation with task-specific edge detection using
  cnns and a discriminatively trained domain transform.
\newblock In {\em Proceedings of the IEEE Conference on Computer Vision and
  Pattern Recognition (CVPR)}, pages 4545--4554, 2016.

\bibitem{kendall2018multi}
Alex Kendall, Yarin Gal, and Roberto Cipolla.
\newblock Multi-task learning using uncertainty to weigh losses for scene
  geometry and semantics.
\newblock In {\em Proceedings of the IEEE Conference on Computer Vision and
  Pattern Recognition (CVPR)}, pages 7482--7491, 2018.

\bibitem{kokkinos2017ubernet}
Iasonas Kokkinos.
\newblock Ubernet: Training a universal convolutional neural network for low-,
  mid-, and high-level vision using diverse datasets and limited memory.
\newblock In {\em Proceedings of the IEEE Conference on Computer Vision and
  Pattern Recognition (CVPR)}, pages 6129--6138, 2017.

\bibitem{xu2018pad}
Dan Xu, Wanli Ouyang, Xiaogang Wang, and Nicu Sebe.
\newblock Pad-net: Multi-tasks guided prediction-and-distillation network for
  simultaneous depth estimation and scene parsing.
\newblock In {\em Proceedings of the IEEE Conference on Computer Vision and
  Pattern Recognition (CVPR)}, pages 675--684, 2018.

\bibitem{standley2020tasks}
Trevor Standley, Amir Zamir, Dawn Chen, Leonidas Guibas, Jitendra Malik, and
  Silvio Savarese.
\newblock Which tasks should be learned together in multi-task learning?
\newblock In {\em International Conference on Machine Learning}, pages
  9120--9132. PMLR, 2020.

\bibitem{nakano2021cross}
Akihiro Nakano, Shi Chen, and Kazuyuki Demachi.
\newblock Cross-task consistency learning framework for multi-task learning.
\newblock {\em arXiv preprint arXiv:2111.14122}, 2021.

\bibitem{zhu2020edge}
Shengjie Zhu, Garrick Brazil, and Xiaoming Liu.
\newblock The edge of depth: Explicit constraints between segmentation and
  depth.
\newblock In {\em Proceedings of the IEEE Conference on Computer Vision and
  Pattern Recognition (CVPR)}, pages 13116--13125, 2020.

\bibitem{papadopoulos2021semantic}
Sotirios Papadopoulos, Ioannis Mademlis, and Ioannis Pitas.
\newblock Semantic image segmentation guided by scene geometry.
\newblock In {\em 2021 IEEE International Conference on Autonomous Systems
  (ICAS)}, pages 1--5. IEEE, 2021.

\bibitem{paszke2016enet}
Adam Paszke, Abhishek Chaurasia, Sangpil Kim, and Eugenio Culurciello.
\newblock Enet: A deep neural network architecture for real-time semantic
  segmentation.
\newblock {\em arXiv preprint arXiv:1606.02147}, 2016.

\bibitem{mehta2018espnet}
Sachin Mehta, Mohammad Rastegari, Anat Caspi, Linda Shapiro, and Hannaneh
  Hajishirzi.
\newblock Espnet: Efficient spatial pyramid of dilated convolutions for
  semantic segmentation.
\newblock In {\em Proceedings of the european conference on computer vision
  (ECCV)}, pages 552--568, 2018.

\bibitem{zhang2018shufflenet}
Xiangyu Zhang, Xinyu Zhou, Mengxiao Lin, and Jian Sun.
\newblock Shufflenet: An extremely efficient convolutional neural network for
  mobile devices.
\newblock In {\em Proceedings of the IEEE conference on computer vision and
  pattern recognition (CVPR)}, pages 6848--6856, 2018.

\bibitem{sandler2018mobilenetv2}
Mark Sandler, Andrew Howard, Menglong Zhu, Andrey Zhmoginov, and Liang-Chieh
  Chen.
\newblock Mobilenetv2: Inverted residuals and linear bottlenecks.
\newblock In {\em Proceedings of the IEEE conference on computer vision and
  pattern recognition (CVPR)}, pages 4510--4520, 2018.

\bibitem{hinton2015distilling}
Geoffrey Hinton, Oriol Vinyals, and Jeff Dean.
\newblock Distilling the knowledge in a neural network.
\newblock {\em arXiv preprint arXiv:1503.02531}, 2015.

\bibitem{liu2019structured}
Yifan Liu, Ke~Chen, Chris Liu, Zengchang Qin, Zhenbo Luo, and Jingdong Wang.
\newblock Structured knowledge distillation for semantic segmentation.
\newblock In {\em Proceedings of the IEEE conference on computer vision and
  pattern recognition (CVPR)}, pages 2604--2613, 2019.

\bibitem{he2019knowledge}
Tong He, Chunhua Shen, Zhi Tian, Dong Gong, Changming Sun, and Youliang Yan.
\newblock Knowledge adaptation for efficient semantic segmentation.
\newblock In {\em Proceedings of the IEEE Conference on Computer Vision and
  Pattern Recognition (CVPR)}, pages 578--587, 2019.

\bibitem{liu2017learning}
Zhuang Liu, Jianguo Li, Zhiqiang Shen, Gao Huang, Shoumeng Yan, and Changshui
  Zhang.
\newblock Learning efficient convolutional networks through network slimming.
\newblock In {\em Proceedings of the IEEE international conference on computer
  vision (ICCV)}, pages 2736--2744, 2017.

\bibitem{jacob2018quantization}
Benoit Jacob, Skirmantas Kligys, Bo~Chen, Menglong Zhu, Matthew Tang, Andrew
  Howard, Hartwig Adam, and Dmitry Kalenichenko.
\newblock Quantization and training of neural networks for efficient
  integer-arithmetic-only inference.
\newblock In {\em Proceedings of the IEEE conference on computer vision and
  pattern recognition (CVPR)}, pages 2704--2713, 2018.

\bibitem{sun2017learning}
Haoran Sun, Xiangyi Chen, Qingjiang Shi, Mingyi Hong, Xiao Fu, and Nikos~D
  Sidiropoulos.
\newblock Learning to optimize: Training deep neural networks for wireless
  resource management.
\newblock In {\em 2017 IEEE 18th International Workshop on Signal Processing
  Advances in Wireless Communications (SPAWC)}, pages 1--6. IEEE, 2017.

\bibitem{hong2015decoupled}
Seunghoon Hong, Hyeonwoo Noh, and Bohyung Han.
\newblock Decoupled deep neural network for semi-supervised semantic
  segmentation.
\newblock {\em Advances in neural information processing systems}, 28, 2015.

\bibitem{hu2020real}
Ping Hu, Federico Perazzi, Fabian~Caba Heilbron, Oliver Wang, Zhe Lin, Kate
  Saenko, and Stan Sclaroff.
\newblock Real-time semantic segmentation with fast attention.
\newblock {\em IEEE Robotics and Automation Letters}, 6(1):263--270, 2020.

\bibitem{xie2015holistically}
Saining Xie and Zhuowen Tu.
\newblock Holistically-nested edge detection.
\newblock In {\em Proceedings of the IEEE international conference on computer
  vision (ICCV)}, pages 1395--1403, 2015.

\bibitem{yu2017casenet}
Zhiding Yu, Chen Feng, Ming-Yu Liu, and Srikumar Ramalingam.
\newblock Casenet: Deep category-aware semantic edge detection.
\newblock In {\em Proceedings of the IEEE Conference on Computer Vision and
  Pattern Recognition (CVPR)}, pages 5964--5973, 2017.

\bibitem{hu2019dynamic}
Yuan Hu, Yunpeng Chen, Xiang Li, and Jiashi Feng.
\newblock Dynamic feature fusion for semantic edge detection.
\newblock {\em arXiv preprint arXiv:1902.09104}, 2019.

\bibitem{lin2017feature}
Tsung-Yi Lin, Piotr Doll{\'a}r, Ross Girshick, Kaiming He, Bharath Hariharan,
  and Serge Belongie.
\newblock Feature pyramid networks for object detection.
\newblock In {\em Proceedings of the IEEE Conference on Computer Vision and
  Pattern Recognition (CVPR)}, pages 2117--2125, 2017.

\bibitem{canny1986computational}
John Canny.
\newblock A computational approach to edge detection.
\newblock {\em IEEE Transactions on pattern analysis and machine intelligence},
  (6):679--698, 1986.

\bibitem{liu2021paddleseg}
Yi~Liu, Lutao Chu, Guowei Chen, Zewu Wu, Zeyu Chen, Baohua Lai, and Yuying Hao.
\newblock Paddleseg: A high-efficient development toolkit for image
  segmentation.
\newblock {\em arXiv preprint arXiv:2101.06175}, 2021.

\bibitem{cordts2016cityscapes}
Marius Cordts, Mohamed Omran, Sebastian Ramos, Timo Rehfeld, Markus Enzweiler,
  Rodrigo Benenson, Uwe Franke, Stefan Roth, and Bernt Schiele.
\newblock The cityscapes dataset for semantic urban scene understanding.
\newblock In {\em Proceedings of the IEEE Conference on Computer Vision and
  Pattern Recognition (CVPR)}, pages 3213--3223, 2016.

\bibitem{rahnemoonfar2021floodnet}
Maryam Rahnemoonfar, Tashnim Chowdhury, Argho Sarkar, Debvrat Varshney, Masoud
  Yari, and Robin~Roberson Murphy.
\newblock Floodnet: A high resolution aerial imagery dataset for post flood
  scene understanding.
\newblock {\em IEEE Access}, 9:89644--89654, 2021.

\bibitem{he2016deep}
Kaiming He, Xiangyu Zhang, Shaoqing Ren, and Jian Sun.
\newblock Deep residual learning for image recognition.
\newblock In {\em Proceedings of the IEEE Conference on Computer Vision and
  Pattern Recognition (CVPR)}, pages 770--778, 2016.

\bibitem{ma2019paddlepaddle}
Yanjun Ma, Dianhai Yu, Tian Wu, and Haifeng Wang.
\newblock Paddlepaddle: An open-source deep learning platform from industrial
  practice.
\newblock {\em Frontiers of Data and Domputing}, 1(1):105--115, 2019.

\bibitem{zhao2018icnet}
Hengshuang Zhao, Xiaojuan Qi, Xiaoyong Shen, Jianping Shi, and Jiaya Jia.
\newblock Icnet for real-time semantic segmentation on high-resolution images.
\newblock In {\em Proceedings of the European conference on computer vision
  (ECCV)}, pages 405--420, 2018.

\bibitem{mehta2019espnetv2}
Sachin Mehta, Mohammad Rastegari, Linda Shapiro, and Hannaneh Hajishirzi.
\newblock Espnetv2: A light-weight, power efficient, and general purpose
  convolutional neural network.
\newblock In {\em Proceedings of the IEEE Conference on Computer Vision and
  Pattern Recognition (CVPR)}, pages 9190--9200, 2019.

\bibitem{wang2021swiftnet}
Haochen Wang, Xiaolong Jiang, Haibing Ren, Yao Hu, and Song Bai.
\newblock Swiftnet: Real-time video object segmentation.
\newblock In {\em Proceedings of the IEEE Conference on Computer Vision and
  Pattern Recognition (CVPR)}, pages 1296--1305, 2021.

\bibitem{yu2018bisenet}
Changqian Yu, Jingbo Wang, Chao Peng, Changxin Gao, Gang Yu, and Nong Sang.
\newblock Bisenet: Bilateral segmentation network for real-time semantic
  segmentation.
\newblock In {\em Proceedings of the European conference on computer vision
  (ECCV)}, pages 325--341, 2018.

\bibitem{yu2021bisenet}
Changqian Yu, Changxin Gao, Jingbo Wang, Gang Yu, Chunhua Shen, and Nong Sang.
\newblock Bisenet v2: Bilateral network with guided aggregation for real-time
  semantic segmentation.
\newblock {\em International Journal of Computer Vision}, 129(11):3051--3068,
  2021.

\bibitem{chen2019fasterseg}
Wuyang Chen, Xinyu Gong, Xianming Liu, Qian Zhang, Yuan Li, and Zhangyang Wang.
\newblock Fasterseg: Searching for faster real-time semantic segmentation.
\newblock {\em arXiv preprint arXiv:1912.10917}, 2019.

\bibitem{fan2021rethinking}
Mingyuan Fan, Shenqi Lai, Junshi Huang, Xiaoming Wei, Zhenhua Chai, Junfeng
  Luo, and Xiaolin Wei.
\newblock Rethinking bisenet for real-time semantic segmentation.
\newblock In {\em Proceedings of the IEEE Conference on Computer Vision and
  Pattern Recognition (CVPR)}, pages 9716--9725, 2021.

\bibitem{peng2022pp}
Juncai Peng, Yi~Liu, Shiyu Tang, Yuying Hao, Lutao Chu, Guowei Chen, Zewu Wu,
  Zeyu Chen, Zhiliang Yu, Yuning Du, et~al.
\newblock Pp-liteseg: A superior real-time semantic segmentation model.
\newblock {\em arXiv preprint arXiv:2204.02681}, 2022.

\bibitem{holder2022efficient}
Christopher~J Holder and Muhammad Shafique.
\newblock On efficient real-time semantic segmentation: A survey.
\newblock {\em arXiv preprint arXiv:2206.08605}, 2022.

\bibitem{kong2018recurrent}
Shu Kong and Charless~C Fowlkes.
\newblock Recurrent scene parsing with perspective understanding in the loop.
\newblock In {\em Proceedings of the IEEE Conference on Computer Vision and
  Pattern Recognition (CVPR)}, pages 956--965, 2018.

\bibitem{wang2022active}
Chi Wang, Yunke Zhang, Miaomiao Cui, Peiran Ren, Yin Yang, Xuansong Xie,
  Xian-Sheng Hua, Hujun Bao, and Weiwei Xu.
\newblock Active boundary loss for semantic segmentation.
\newblock In {\em Proceedings of the AAAI Conference on Artificial
  Intelligence}, volume~36, pages 2397--2405, 2022.

\bibitem{loshchilov2016sgdr}
Ilya Loshchilov and Frank Hutter.
\newblock Sgdr: Stochastic gradient descent with warm restarts.
\newblock {\em arXiv preprint arXiv:1608.03983}, 2016.

\bibitem{chen2018encoder}
Liang-Chieh Chen, Yukun Zhu, George Papandreou, Florian Schroff, and Hartwig
  Adam.
\newblock Encoder-decoder with atrous separable convolution for semantic image
  segmentation.
\newblock In {\em Proceedings of the European conference on computer vision
  (ECCV)}, pages 801--818, 2018.

\end{thebibliography}
\end{document}